\documentclass[10pt,twocolumn,letterpaper]{article}

\usepackage[pagenumbers]{cvpr} 

\usepackage{float}
\usepackage{graphicx}
\usepackage{amsmath}
\usepackage{amssymb}
\usepackage{booktabs}
\usepackage{xcolor}

\definecolor{cvprblue}{rgb}{0.21,0.49,0.74}
\usepackage[pagebackref,breaklinks,colorlinks,citecolor=cvprblue]{hyperref}

\def\paperID{1241} 
\def\confName{CVPR}
\def\confYear{2024}

\title{Weakly-Supervised 3D Reconstruction of Clothed Humans via Normal Maps}

\author{
Jane Wu\textsuperscript{1} \qquad Diego Thomas\textsuperscript{2} \qquad Ronald Fedkiw\textsuperscript{1,3} \\
\textsuperscript{1}Stanford University \qquad \textsuperscript{2}Kyushu University \qquad \textsuperscript{3}Epic Games\\
{\tt\small janehwu@stanford.edu,thomas@ait.kyushu-u.ac.jp,fedkiw@cs.stanford.edu}
}

\begin{document}
\maketitle

\begin{abstract}
We present a novel deep learning-based approach to the 3D reconstruction of clothed humans using weak supervision via 2D normal maps.
Given a single RGB image or multiview images, our network infers a signed distance function (SDF) discretized on a tetrahedral mesh surrounding the body in a rest pose.
Subsequently, inferred pose and camera parameters are used to generate a normal map from the SDF.
A key aspect of our approach is the use of Marching Tetrahedra to (uniquely) compute a triangulated surface from the SDF on the tetrahedral mesh, facilitating straightforward differentiation (and thus backpropagation).
Thus, given only ground truth normal maps (with no volumetric information ground truth information), we can train the network to produce SDF values from corresponding RGB images.
Optionally, an additional multiview loss leads to improved results.
We demonstrate the efficacy of our approach for both network inference and 3D reconstruction.
\end{abstract}

\section{Introduction}
Recent work on 3D human digitization has largely focused on the fully-supervised setting, where deep neural networks (DNNs) are trained to explicitly fit so-called ground truth 3D geometry \cite{saito2020pifuhd,zheng2021pamir,xiu2022icon,xiu2023econ}.
In such approaches, high-end capture setups (with 4D scanners or a large number of cameras) are typically used to obtain high-quality, multiview training data \cite{pons2017clothcap,xiang2021modeling}.
Inferring 3D geometry and appearance from 2D information is a highly underconstrained problem; thus, it can be challenging for models trained on such high-quality data to generalize to the lower quality images typical of consumer-grade devices (such as phones and webcams).
However, the ability to do so is crucial to the democratization of digital humans required for many applications in AR/VR, robotics, healthcare, etc.

\begin{figure}[ht]
    \centering
    \includegraphics[width=\linewidth]{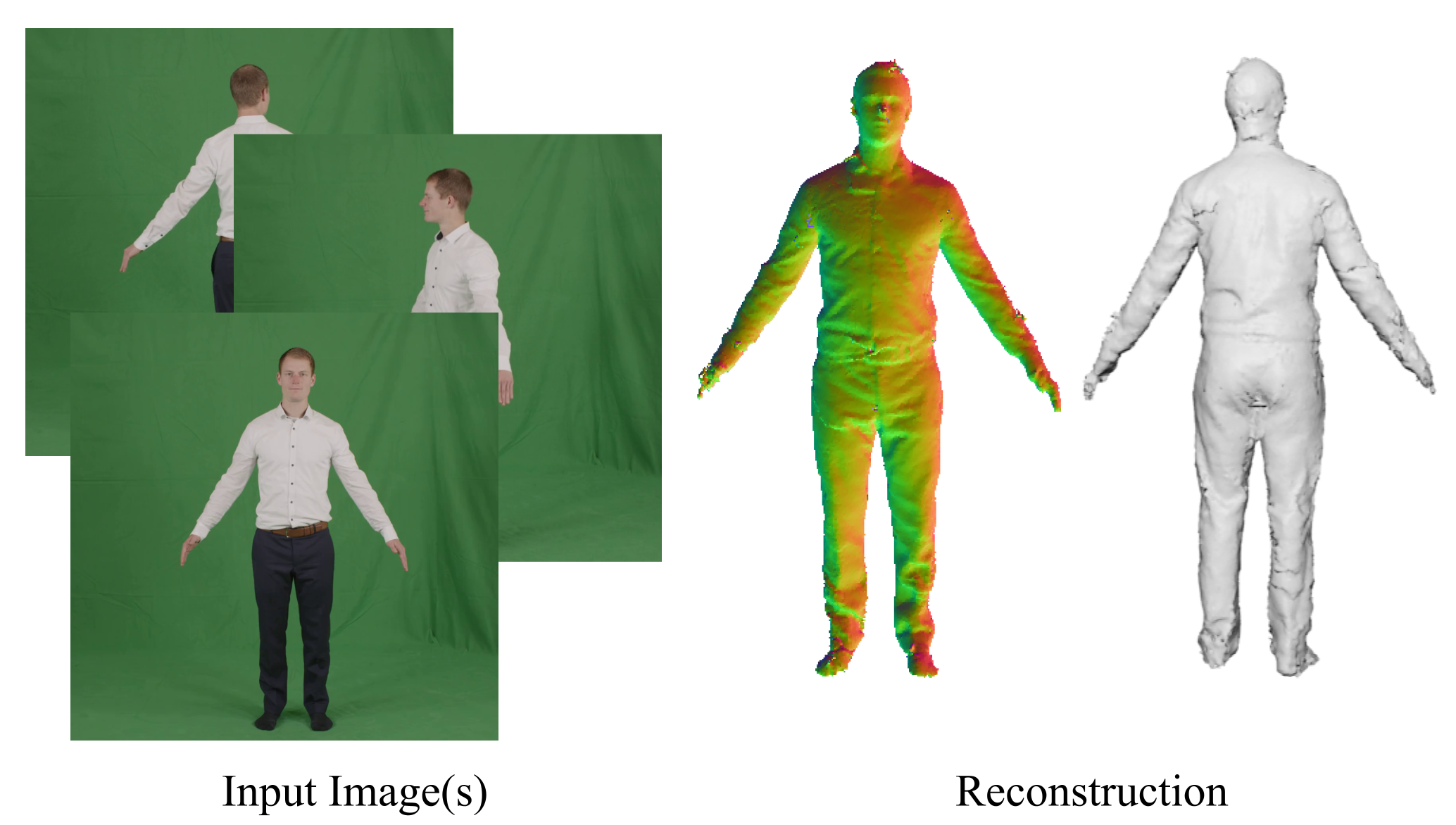}
    \caption{Given in-the-wild image(s) or video, our method is able to reconstruct clothed humans using inferred normal maps as the supervisory signal. Sample frames from the input video are shown to the left, and the predicted triangle mesh is shown to the right (the front-facing mesh is shaded with its normal map).}
    \label{fig:teaser}
\end{figure}

\begin{figure*}[t]
    \centering
    \includegraphics[width=\linewidth]{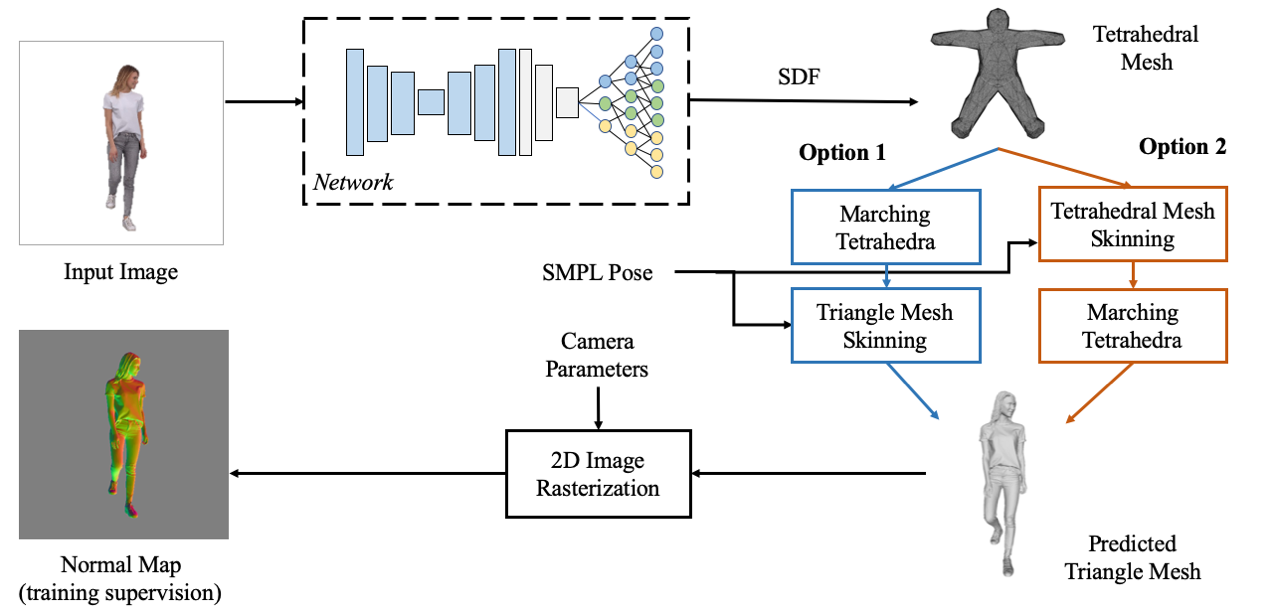}
    \caption{Given an RGB image, a graph-based DNN infers SDF values on a tetrahedral mesh parameterizing a volume surrounding/containing the body. Option 1: Marching Tetrahedra is used to compute a triangle mesh from this implicit surface, and the resulting mesh is skinned to an estimated pose. Option 2: Alternatively, the tetrahedral mesh can be skinned first before using Marching Tetrahedra. During training (only), a normal map (corresponding to inferenced pose and camera parameters) is generated for the predicted clothed body mesh and compared with the ground truth.}
    \label{fig:pipeline}
\end{figure*}

Given a single (monocular) RGB image, we estimate the clothed body with a DNN that infers signed distance function (SDF) values for each vertex of a tetrahedral mesh surrounding the body (similar to \cite{lee2018skinned,lee2019robust,onizuka2020tetratsdf,wu2020skinning}).
Implicit surfaces have become a common choice for 3D reconstruction from images (see e.g.\ \cite{saito2020pifuhd,zheng2021pamir,peng2021neural,tiwari2021neural,chen2022gdna,xiu2022icon,xiu2023econ}), especially as neural radiance fields (NeRFs) \cite{mildenhall2020nerf} have gained popularity.
Importantly, the tetrahedral mesh data structure enables our use of Marching Tetrahedra to uniquely compute a triangulated surface.
The resulting algorithm is straightforward to differentiate, alleviating concerns associated with the nondeterministic nature of Marching Cubes (see e.g.\ \cite{lorensen1987marching,remelli2020meshsdf}) or the ray tracing of an implicit surface (see e.g.\ \cite{niemeyer2020differentiable,yariv2020multiview,chen2022gdna,vicini2022differentiable}).

Our approach is a follow-up to \cite{onizuka2020tetratsdf}, which also uses an explicit representation of an SDF on a tetrahedral mesh; however, we add a second explicit representation of the surface via a triangle mesh.
We thus have access to two explicit versions of the neural SDF, and energies can be conveniently formulated for either the volume or the surface or both.
While similar in spirit to \cite{mehta2022level}, our method does not require the construction of a velocity in order to capture these energies with an evolving level set function; thus, we can control mesh based properties (e.g. \ area and dihedral angles) that would be lost when converting to a velocity field.
On the other hand, the approach in \cite{mehta2022level} could be used to alleviate locking concerns in cases where discretizations on the triangle mesh and the tetrahedral mesh do not interact as expected due to differences in the degrees of freedom.

The main goal of our work is to provide weak supervision during DNN training via 2D normal maps \cite{gabeur2019moulding,natsume2019siclope,smith2019facsimile,saito2020pifuhd,chen2022gdna,xiu2022icon,xiu2023econ,jiang2022selfrecon}.
By formulating the optimization problem with respect to image-based normals, we aim to better represent fundamental correlations between 2D images and 3D geometry in order to facilitate subsequent democratization to consumer-grade devices.
Given (inferenced) pose and camera parameters, a normal map can be computed from the skinned triangulated surface.
See Figure \ref{fig:pipeline}.
While we initialize the SDF network parameters with the data from \cite{onizuka2020tetratsdf}, the resulting model will tend to overfit (since it is trained on a limited quantity of 3D data) and thus generalize poorly to in-the-wild 2D images.
 Thus, after this initialization, we train the model in a weakly supervised manner using only ground truth normal maps.

To summarize, our contributions:
\begin{itemize}
\item We illustrate that our network can be used to reconstruct 3D geometry from sparse multiview RGB data obtained with consumer-grade cameras (and no ground truth 3D labels).
\item We compute the correct gradients to differentiate through Marching Tetrahedra via a Lagrangian formulation, which enables differentiable mesh generation and thus end-to-end training with both volumetric and surface-based energies.
\item We present a differentiable image rasterizer that: (1) allows us to use normal maps for weak supervision and (2) can efficiently compute normal maps from triangle meshes with over 300k triangles during network training.
\item We formulate regularization energies that coerce inferred implicit surfaces to: (1) resemble true SDFs and (2) be locally smooth.
\item We formulate silhouette energies defined to enforce 3D boundary matching.
\end{itemize}

\label{sec:intro}

\section{Related Work}
\subsection{Human Shape Estimation}
Various works use parametric body models such as SMPL \cite{loper2015smpl} to estimate human body pose and shape without clothing \cite{liu2022recent,goel2023humans,kim2023sampling,nam2023cyclic}.
While existing methods are able to generalize to in-the-wild images, the inferred body mesh is often quite different from the underlying body shape and does not capture clothing.

In order to reconstruct humans wearing clothing from a single image \cite{tian2023recovering}, template-based approaches either rely on parametric models \cite{kolotouros2021probabilistic,corona2022structured,jinka2023sharp,zhang2023global,zheng2023avatarrex} or use person-specific meshes \cite{yu2019simulcap,habermann2020deepcap}.
For instance, GTA \cite{zhang2023global} projects SMPL onto a learned 3D triplane representation, and \cite{zheng2023avatarrex} constructs local implicit fields centered around locations on the SMPL-X model \cite{pavlakos2019expressive}.
Limitations of template-based approaches to clothed human reconstruction include the output being constrained by the topology of the template as well as a reliance on accurate pose estimation.
Template-free methods typically leverage 2D signals or 3D geometric representations to recover geometry.
In \cite{gabeur2019moulding}, reconstruction is achieved by generating front and back depth images that are later combined into a 3D surface.
\cite{han2023high} builds on this idea and proposes a coarse-to-fine reconstruction method leveraging both predicted depth and normal images.
Inspired by shape-from-silhouette techniques, SiCloPe \cite{natsume2019siclope} recovers geometry by predicting silhouette images and 3D joint positions.
\cite{varol2018bodynet,zheng2019deephuman} predict volumetric occupancy on a uniform voxel grid directly, while \cite{feng2022fof} proposed learning a Fourier subspace of 3D occupancy; in both cases, Marching Cubes can be used to generate a triangle mesh.
PIFu \cite{saito2019pifu} and PIFuHD \cite{saito2020pifuhd} infer 3D shape with neural implicit functions sampled onto a grid.
Follow-up work \cite{chan2022integratedpifu,xiu2022icon} leverages predicted normal maps to improve depth inference.
PAMIR \cite{zheng2021pamir} extends PIFuHD to increase generalizability by regularizing the implicit function using semantic features from a parametric model.
ICON \cite{xiu2022icon} and ECON \cite{xiu2023econ} leverage inferred front and back normal maps as an intermediate encoding of 3D geometry, but these methods still rely on ground truth 3D scan data during training.

Instead of a single input image, other works aim to construct animatable avatars from a sparse set of cameras \cite{hong2021stereopifu,shao2022diffustereo,zhao2022humannerf,zhou2022hdhuman}, video \cite{pang2021few,dong2022totalselfscan,feng2022capturing,jiang2022neuman,jiang2022selfrecon,te2022neural,weng2022humannerf,guo2023vid2avatar,jiang2023instantavatar,yu2023monohuman}, depth \cite{wang2021metaavatar,dong2022geometry,xue2023nsf,zheng2023learning}, point clouds \cite{ma2021power,kim2022laplacianfusion}, 4D capture \cite{icsik2023humanrf}, or scans \cite{li2022avatarcap,lin2022learning,shen2023x}.
Most similar to our work, SelfRecon \cite{jiang2022selfrecon} uses normal maps inferred from PIFuHD \cite{saito2020pifuhd} to supervise network training, and SeSDF \cite{cao2023sesdf} can either take as input a single image or uncalibrated multiview images

Moving towards improved generalization, weakly supervised methods have also been explored for human pose estimation \cite{kanazawa2018end,omran2018neural,pavlakos2018learning,yang20183d}, human body shape \cite{zanfir2020weakly,moon20223d}, and garment template reconstruction (on the SMPL body) \cite{moon20223d,de2023drapenet}.


\subsection{Differentiable Marching Cubes / Tetrahedra}\label{sec:related_mt}
A number of recent works have proposed methods for backpropagating through Marching Cubes \cite{lorensen1987marching} / Marching Tetrahedra \cite{treece1999regularised} by either leveraging properties of a point-to-SDF network \cite{remelli2020meshsdf,shen2021deep} (e.g.\ as in DeepSDF \cite{park2019deepsdf}) or by training a DNN for mesh generation \cite{liao2018deep}.
MeshSDF \cite{remelli2020meshsdf} builds on DeepSDF \cite{park2019deepsdf}, where a network $f_{\eta}$ is trained to infer an SDF value at a location $x$ conditioned on a latent shape code $\eta$.
In the forward step, $f_{\eta}(x)$ is computed for every vertex of a fixed voxel grid so that Marching Cubes can be used to generate a triangle mesh.
The authors postulate that a small increase in the SDF values would move a triangle vertex in the normal direction, which is only true idealistically when there are no shocks/rarefactions in the SDF isocontours (see e.g.\ \cite{osher2004level}); moreover, Marching Cubes does not move vertices in such a manner, even when it produces a consistent set of vertices under perturbation.
Their assumptions also necessitate $f_{\eta}$ being a true SDF, even though it is only an approximation. 
Follow-up work in \cite{shen2021deep} presents a similar formulation using Marching Tetrahedra.
See \cite{mehta2022level} for more discussion on the problematic assumptions in \cite{remelli2020meshsdf}.

\label{sec:RW}

\section{Weak Supervision via Normal Maps}
In the context of human digitization, it can be challenging to train generalizable ML-based models with full 3D supervision (see e.g.\ \cite{saito2019pifu,onizuka2020tetratsdf,saito2020pifuhd,zheng2021pamir}).
Robust generalization typically necessitates access to a large number of ground truth training examples; however, publicly available datasets of 3D scan data for clothed human bodies remain scarce (in part, because it is both expensive and complicated to obtain).
Even if such data were more readily available, existing works typically require unskinning the data to a reference pose (see e.g.\ \cite{habermann2019livecap,habermann2020deepcap,onizuka2020tetratsdf,wu2020skinning}), which can lead to various complications: tangling, self-intersection, inversion, etc.

To alleviate dependency on labeled 3D data, we propose a weakly supervised approach using 2D normal maps as ground truth labels (only) during training.
A 2D normal map defines an RGB value for each pixel, corresponding to the (camera or world space) unit normal that best represents the geometry rasterized to that pixel.
A normal map can be approximated by casting a ray through the pixel center and subsequently interpolating normals to the ray-geometry intersection point, although a better estimate would be obtained by supersampling (similar to the way pixel color is computed).
Importantly, difficult to handle occluded regions (such as the armpit) may be ignored (in contrast to full 3D supervision).
Since there are a number of ways to obtain ground truth normal maps (besides utilizing 3D scan data), this approach vastly increasing the amount of data available for training.
For example, one can utilize RGBD images \cite{andriluka20142d,godard2019digging,zhao2020monocular,jun2021monocular}, stereo pairs \cite{badki2020bi3d,kusupati2020normal}, and/or neural networks (including NeRFs \cite{mildenhall2020nerf}) trained to produce normal maps from RGB images \cite{saito2020pifuhd,xiu2022icon,xiu2023econ}.
Increasing the amount of training data (in this way) facilitates generalization to a much more representative and diverse set of people in clothing (as compared to using only a limited number of 3D scans).

Given inferred SDF values $\hat{\phi}_k$ on the (fixed) tetrahedral mesh vertices $u_k$, Marching Tetrahedra is used to uniquely generate a triangle mesh with vertices $v_i(\hat{\phi}_k)$.
Given an inferred pose $\hat{\theta}$ and camera parameters $\hat{c}$, a normal map $N(v_i(\hat{\phi}_k), \hat{\theta}, \hat{c})$ can be generated.
The objective function to be minimized is then
\begin{equation}\label{eq:normal_loss}
    \mathcal{L}(\hat{\phi}_k, \hat{\theta}, \hat{c}) = \left\lVert  N(v_i(\hat{\phi}_k), \hat{\theta}, \hat{c}) - N_{GT}\right\rVert
\end{equation}
where $N_{GT}$ is a ground truth normal map.

\section{Tetrahedral Mesh Framework}\label{sec:tet_mesh}
\begin{figure}[ht]
    \centering
    \includegraphics[width=\linewidth]{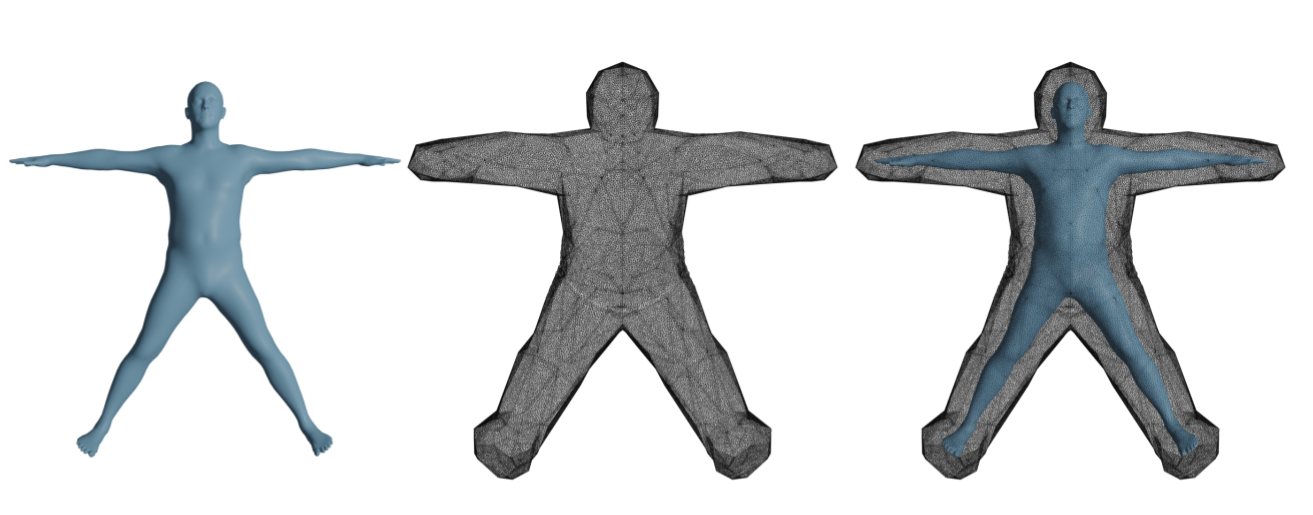}
    \caption{The tetrahedral mesh parameterizes a volume of air surrounding and including the body.}
    \label{fig:outer_shell}
\end{figure}

The 3D space surrounding and including the human body is parameterized via a tetrahedral mesh.
First, a Cartesian grid based level set representation is generated for the SMPL template body \cite{loper2015smpl} in the star pose (similar to \cite{onizuka2020tetratsdf,wu2020skinning}).
Then, a constant value is subtracted from the SDF values in order to inflate the zero level set so that its interior can contain a wide range of clothed body shapes.
Subsequently, a tetrahedral mesh is generated for this interior region using red/green refinement \cite{molino2003crystalline,teran2005adaptive}.
See Figure \ref{fig:outer_shell}.

Given an input image, a DNN is trained to infer an implicit surface approximation to the clothed body, represented by SDF values $\hat\phi_k$ on the tetrahedral mesh vertices $u_k$ (similar to \cite{onizuka2020tetratsdf}).
The DNN is composed of a CNN-based stacked hourglass encoder, followed by graph convolutional layers that progressively increase the resolution of the sampled SDF.
This encodes an input image into a feature vector, which is then decoded into SDF values on the vertices of the tetrahedral mesh.
Due to the large number of tetrahedral mesh vertices, the graph convolutional layers are only partially connected in order to significantly reduce memory usage.

\subsection{Skinning}\label{sec:skinning}
The tetrahedral mesh can be deformed via linear blend skinning (LBS) using per-tetrahedron-vertex, per-joint skinning weights $w_{kj}$.
The skinning weights are assigned in the star pose by first finding the point on the SMPL template body mesh closest to each tetrahedral mesh vertex, and then barycentrically interpolating skinning weights to that point from the vertices of the SMPL template body mesh triangle that contains it.

Given pose parameters $\theta$ with joint transformations $T_j(\theta)$, the skinned position of each tetrahedral mesh vertex is $u_k(\theta) = \sum_j w_{kj}T_j(\theta)u_k^j$
where $u_k^j$ is the location of $u_k$ in the untransformed reference space of joint $j$.
VIBE \cite{kocabas2020vibe} is used to estimate the SMPL pose parameters $\hat{\theta} \in \mathbb{R}^{72}$ for any given input image.
During training, all layers of a pretrained VIBE model are frozen except for the final Gated Recurrent Unit (GRU) layer.

\section{Marching Tetrahedra}\label{sec:mt}

Given SDF values $\phi_k$ defined on tetrahedral mesh vertices $u_k$, Marching Tetrahedra can be implemented to compute a unique (non-ambiguous) triangle mesh with vertices $v_i$, making differentiation more straightforward as compared to the many cases (and non-uniqueness) that need to be considered for Marching Cubes.
In order to avoid triangle vertices $v_i$ coincident with a tetrahedron vertex $u_k$, all the $\phi_k$ values are preprocessed (infinitesimally) changing those with $|\phi_k| < \epsilon$ to $\phi_k = \epsilon\hspace{1mm}\text{sign}(\phi_k)$, e.g.\ with $\epsilon=10^{-8}$.

For each tetrahedron mesh edge $e_i=\{u_{k_1}, u_{k_2}\}$ that includes a sign change, i.e.\ $\text{sign}(\phi_{k_1}) \neq \text{sign}(\phi_{k_2})$, a triangle vertex
\begin{align}\label{eq:zero_crossing}
\begin{split}
  v_i {}& = \frac{-\phi_{k_2}}{\phi_{k_1} - \phi_{k_2}}u_{k_1} + \frac{\phi_{k_1}}{\phi_{k_1} - \phi_{k_2}}u_{k_2} \\
\end{split}
\end{align}
is defined using linear interpolation.
Afterwards, triangles are constructed in a tetrahedron-by-tetrahedron manner by considering the two cases that can occur: either three edges of the tetrahedron contain triangle vertices and one triangle is constructed, or four edges contain triangle vertices and a quadrilateral is constructed and split into two triangles.
Note that this typically arbitrary splitting of the quadrilateral can be made consistent for the sake of differentiation.
Since the tetrahedral mesh does not change topology, the edges can be numbered in a fixed manner; then, one can consistently split a quadrilateral by connecting the triangle vertex on the lowest numbered edge to the triangle vertex on the highest numbered edge (or via a similar alternative strategy).
The resulting triangle mesh is guaranteed to be watertight, and the vertices in each triangle are reordered (when necessary) to ensure that all face normals point outwards.

\subsection{Ray-Tracing the Implicit Surface Directly}
As an alternative to Marching Tetrahedra, consider casting a ray to find an intersection point with the implicit surface and subsequently using the normal vector defined (directly) by the implicit surface at that intersection point.
A number of existing works consider such approaches in various ways, see e.g.\ \cite{niemeyer2020differentiable,yariv2020multiview,bangaru2022differentiable,chen2022gdna,vicini2022differentiable}.
Perturbations of the intersection point depend on perturbations of the $\phi$ values on the vertices of the tetrahedron that the intersection point lies within.
If a change in $\phi$ values causes the intersection point to no longer be contained inside the tetrahedron, one would need to discontinuously jump to some other tetrahedron (which could be quite far away, if it even exists).
A potential remedy for this would be to define a virtual implicit surface that extends out of the tetrahedron in a way that provides some sort of continuity (especially along silhouette boundaries).

Comparatively, our Marching Tetrahedra approach allows us to presume (for example) that the point of intersection remains fixed on the face of the triangle even as the triangle moves.
Since the implicit surface has no explicit parameterization, one is unable to similarly hold the intersection point fixed.
The implicit surface utilizes an Eulerian point of view where the rays (which represent the discretization) are held fixed while the implicit surface moves (as $\phi$ values change), in contrast to our Lagrangian discretization where the rays are allowed to move/bend in order to follow fixed intersection points during differentiation.
A similar approach for an implicit surface would hold the intersection point inside the tetrahedron fixed even as $\phi$ changes.
Although such an approach holds potential due to the fact that implicit surfaces are amenable to computing derivatives off of the surface itself, the merging/pinching of isocontours created by convexity/concavity would likely lead to various difficulties.
Furthermore, other issues would need to be addressed as well, e.g.\ the gradients (and thus normals) are only piecewise constant (and thus discontinuous) in the piecewise linear tetrahedral mesh basis.

\subsection{Computing Gradients}\label{sec:mt_grads}
According to Equation \ref{eq:zero_crossing},
\begin{equation}\label{eq:mt_grad}
    \frac{\partial v_i}{\partial(\phi_{k_1}, \phi_{k_2})} 
    = \begin{bmatrix}\dfrac{\phi_{k_2}(u_{k_1}- u_{k_2})}{(\phi_{k_1}-\phi_{k_2})^2} \quad
    \dfrac{\phi_{k_1}(u_{k_2} -u_{k_1})}{(\phi_{k_1}-\phi_{k_2})^2}\end{bmatrix}
\end{equation}
where dividing by $(\phi_{k_1} - \phi_{k_2})^2$ can be problematic.
The preprocess at the beginning of Section \ref{sec:mt} guarantees that $|\phi_{k_1} - \phi_{k_2}| \geq 2 \epsilon$, which means that the worst possible scenario for Equation \ref{eq:zero_crossing} (when $|\phi_{k_1}|=|\phi_{k_2}|=\epsilon$) still results in $\mathcal{O}(1)$ coefficients for $u_{k_1}$ and $u_{k_2}$; however, the $\phi$-based coefficients in Equation \ref{eq:mt_grad} would be $\mathcal{O}(1 / \epsilon)$.
Thus, while $\epsilon=10^{-8}$ is sufficient for Equation \ref{eq:zero_crossing}, a larger value of $\epsilon$ might be prudent when considering Equation \ref{eq:mt_grad}.

\subsection{Skinning}
There are two options for the algorithm ordering between skinning and Marching Tetrahedra (the latter of which reverses the order in Figure \ref{fig:pipeline}).
For skinning the triangle mesh, the skinned position of each triangle mesh vertex is $v_i(\theta,\phi) = \sum_j w_{ij}(\phi)T_j(\theta)v_i^j(\phi)$ where $v_i^j$ is the location of $v_i$ in the untransformed reference space of joint $j$.
Unlike in Section \ref{sec:skinning} where $w_{kj}$ and $u_k^j$ were fixed, $w_{ij}$ and $v_i^j$ both vary yielding three terms in the product rule.
$\partial v_{i}^j/\partial\phi$ is computed according to Equation \ref{eq:mt_grad}, noting that $u_{k_1}$ and $u_{k_2}$ are fixed.
$w_{ij}(\phi)$ is defined similarly to Equation \ref{eq:zero_crossing},
\begin{equation}\label{eq:weights_crossing}
    w_{ij} = \frac{-\phi_{k_2}}{\phi_{k_1} - \phi_{k_2}}w_{k_1j} + \frac{\phi_{k_1}}{\phi_{k_1} - \phi_{k_2}}w_{k_2j}
\end{equation}
where $w_{k_1 j}$ and $w_{k_2 j}$ are fixed; similar to Equation \ref{eq:mt_grad}, $\partial w_{ij} / \partial \phi$ will contain $\mathcal{O}(1/\epsilon)$ coefficients.
For skinning the tetrahedral mesh, Equations \ref{eq:zero_crossing} and \ref{eq:mt_grad} directly define $v_i$ and $\partial v_i/\partial\phi$ since the skinning is moved to the tetrahedral mesh vertices $u_k$.
Then, $\partial v_i/ \partial u_k$ is computed according to Equation \ref{eq:zero_crossing} in order to chain rule to skinning (i.e.\ to $\partial u_k/ \partial \theta$, which is computed according to the equations in Section \ref{sec:skinning}).

\section{Image Rasterization}
Given a skinned triangulated surface and parameters for a perspective camera model, a camera space normal map is computed using a right-handed coordinate system.
We assume that the geometry is centered in the image, since images are cropped and rescaled during preprocessing.
Normal maps made using different assumptions, or decoded and stored as RGB values, are readily transformed back into unit normals (in camera space) in order to match our assumptions.

\subsection{Normals}\label{sec:geometry}
Recall (from Section \ref{sec:mt}) that triangle vertices are reordered (if necessary) in order to obtain outward-pointing face normals.
The area-weighted outward face normal is
\begin{equation}\label{eq:face_normal}
    n_f(v_1,v_2,v_3) =\frac 12 (v_2-v_1) \times (v_3-v_1)
\end{equation}
where
\begin{equation}\label{eq:area_3d}
    Area(v_1,v_2,v_3) = \frac 12||(v_2-v_1) \times (v_3-v_1)||_2
\end{equation}
is the area weighting.
Area-averaged vertex unit normals $\hat{n}_v$ are computed via
\begin{align}\label{eq:vert_normal}
\begin{split}
     n_v = \sum_f n_f \quad \quad  \quad \quad 
     \hat{n}_v &= \frac{n_v}{||n_v||_2}
\end{split}
\end{align}
where $f$ ranges over all the triangle faces that include vertex $v$. Note that one can drop the 1/2 in Equation \ref{eq:face_normal}, since it cancels out when computing $\hat{n}_v$ in Equation \ref{eq:vert_normal}.

\subsection{Camera Model}\label{sec:camera_model}
\begin{figure}[ht]
    \centering
    \includegraphics[width=\linewidth]{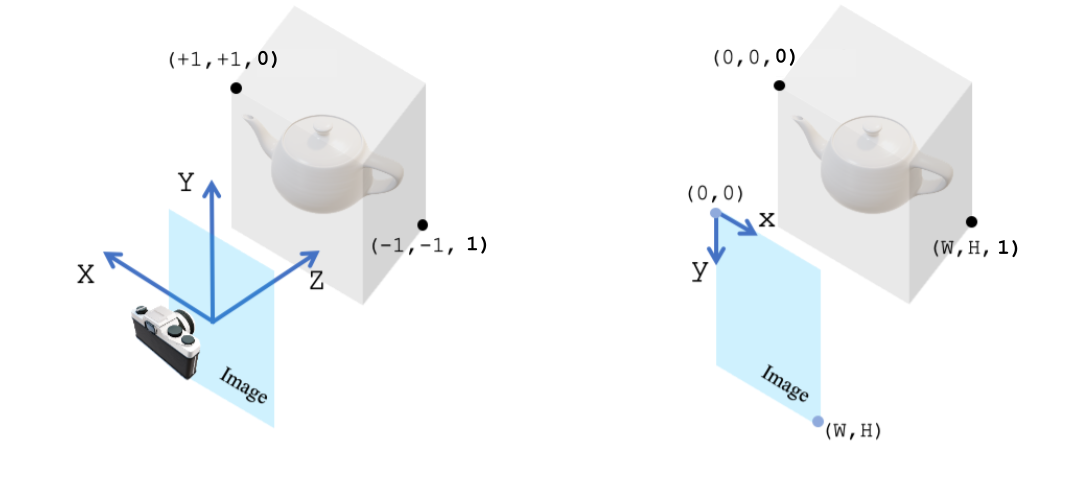}
    \caption[]{The normalized device (left) and screen (right) coordinate systems used during rasterization (based on Pytorch3D conventions\protect\footnotemark).}
    \label{fig:coord_systems}
\end{figure}

The camera rotation and translation are used to transform each vertex $v_g$ of the geometry to the camera view coordinate system (where the origin is located at the camera aperture), i.e.\ $v_{c} = Rv_g + T$.
The normalized device coordinate system normalizes geometry in the viewing frustum (with $z\in [n, f]$) so that all $x,y \in [-1,1]$ and all $z \in [0,1]$.
See Figure \ref{fig:coord_systems}, left.
Vertices are transformed into this coordinate system via
\begin{align}\label{eq:projection_matrix}
\begin{split}
    \begin{bmatrix}[v_{NDC}]\hspace{1mm} z_c\\ z_c\end{bmatrix} &=  \begin{bmatrix}
    \frac{2n}{W} & 0 & 0  & 0 \\
    0 & \frac{2n}{H} & 0 & 0\\
    0 & 0 & \frac{f}{f-n} & \frac{-fn}{f-n} \\
    0 & 0 & 1 & 0 \end{bmatrix}\begin{bmatrix}[v_c]\\1 \end{bmatrix}
\end{split}
\end{align}
where $H = 2n\tan(\theta_{\text{fov}}/2)$ is the height of the image, $\theta_{\text{fov}}$ is the field of view, $W=Ha$ is the width of the image, and $a$ is the aspect ratio.
The screen coordinate system is obtained by transforming the origin to the top left corner of the image, with $+x$ pointing right and $+y$ pointing down.
See Figure \ref{fig:coord_systems}, right.
Vertices are transformed into this coordinate system via
\begin{equation}\label{eq:screen_proj}
    \begin{bmatrix}[v']\\1 \end{bmatrix} = \begin{bmatrix}
    -W/2 & 0 & 0 & W/2 \\
    0 & -H/2 & 0 & H/2\\
    0 & 0 & 1 & 0\\
    0 & 0 & 0 & 1 \\
    \end{bmatrix}\begin{bmatrix}[v_{NDC}]\\1 \end{bmatrix}
\end{equation}

or via
\begin{align}\label{eq:full_transform}
\begin{split}
    \begin{bmatrix}[v']\hspace{1mm} z_c\\ z_c\end{bmatrix} &=
    \begin{bmatrix}
    -n & 0 & W/2 & 0\\
    0 & -n & H/2 & 0\\
    0 & 0 & \frac{f}{f-n} & \frac{-fn}{f-n} \\
    0 & 0 & 1 & 0\\
    \end{bmatrix}
     \begin{bmatrix}[v_c]\\1 \end{bmatrix}\\
\end{split}
\end{align}
which is obtained by multiplying both sides of Equation \ref{eq:screen_proj} by $z_c$ and substituting in Equation \ref{eq:projection_matrix}.

\footnotetext{https://pytorch3d.org/docs/cameras}

\subsection{Normal Map}
For each pixel, a ray is cast from the camera aperture through the pixel center to find its first intersection with the triangulated surface at a point $p$ in world space.
Denoting $v_1, v_2, v_3$ as the vertices of the intersected triangle, barycentric weights for the intersection point
\begin{align}\label{eq:ray_trace}
\begin{split}
    \hat{\alpha}_1 &= \frac{Area(p, v_2, v_3)}{Area(v_1, v_2, v_3)} \\
    \hat{\alpha}_2 &= \frac{Area(v_1, p, v_3)}{Area(v_1, v_2, v_3)}  \\
    \hat{\alpha}_3 &= \frac{Area(v_1, v_2,p)}{Area(v_1, v_2, v_3)}  
\end{split}
\end{align}
are used to compute a rotated (into screen space) unit normal from the unrotated vertex unit normals (see Equation \ref{eq:vert_normal}) via
\begin{equation}\label{eq:normals}
    \hat{n} = R \frac{\hat{\alpha}_1 \hat{n}_{v_1} + \hat{\alpha}_2 \hat{n}_{v_2} + \hat{\alpha}_3 \hat{n}_{v_3}}{||\hat{\alpha}_1 \hat{n}_{v_1} + \hat{\alpha}_2 \hat{n}_{v_2} + \hat{\alpha}_3 \hat{n}_{v_3}||}
\end{equation}
for the normal map.
Note that dropping the denominators in Equation \ref{eq:ray_trace} does not change $\hat{n}$.

\subsection{Scanline Rendering}
After projecting a visible triangle into the screen coordinate system (via Equation \ref{eq:full_transform}), its projected area can be computed as
\begin{align}\label{eq:area_2d}
\begin{split}
    Area2D(v_1', v_2', v_3') &= -\frac 12det\begin{pmatrix}
    x_{2}'-x_{1}' & y_{2}'-y_{1}' \\
    x_{3}'-x_{1}' & y_{3}'-y_{1}' \\
    \end{pmatrix}
\end{split}
\end{align}
similar to Equation \ref{eq:area_3d} (where the negative sign accounts for the fact that visible triangles have normals pointing towards the camera).
When a projected triangle overlaps a pixel center $p'$, barycentric weights for $p'$ are computed by using $Area2D$ instead of $Area$ in Equation \ref{eq:ray_trace}.
Notably, un-normalized world space barycentric weights can be computed from un-normalized screen space barycentric weights via $\alpha_1 = z_2' z_3' \alpha_1'$, $\alpha_2 = z_1' z_3' \alpha_2'$, $\alpha_3 = z_1' z_2'\alpha_3'$ or
\begin{align}\label{eq:alphas}
\begin{split}
    \alpha_1 &= z_2' z_3' Area2D(p', v_2', v_3') \\
    \alpha_2 &= z_1' z_3' Area2D(v_1', p', v_3') \\
    \alpha_3 &= z_1' z_2' Area2D(v_1', v_2', p') 
\end{split}
\end{align}
giving 
\begin{equation}\label{eq:new_normals}
    \hat{n} = R\frac{\alpha_1 \hat{n}_{v_1} + \alpha_2 \hat{n}_{v_2} + \alpha_3 \hat{n}_{v_3}}{||\alpha_1 \hat{n}_{v_1} + \alpha_2 \hat{n}_{v_2} + \alpha_3 \hat{n}_{v_3}||}
\end{equation}
as an (efficient) alternative to Equation \ref{eq:normals}.
If more than one triangle overlaps $p'$, the closest one (i.e.\ the one with the smallest value of $z' = \hat{\alpha}_1'z_1'+\hat{\alpha}_2'z_2'+\hat{\alpha}_3'z_3'$ at $p'$) is chosen.

\subsection{Computing Gradients}
For each pixel overlapped by the triangle mesh, the derivative of the normal (in Equation \ref{eq:new_normals}) with respect to the vertices of the triangle mesh is required, i.e.\ $\partial \alpha_i / \partial v_g$ and $\partial \hat{n}_{v_i} / \partial v_g$ are required.
$\partial \alpha_i / \partial v'$ can be computed from Equations \ref{eq:alphas} and \ref{eq:area_2d}, $\partial v'/\partial v_c$ can be computed from Equation \ref{eq:full_transform}, and $\partial v_c/\partial v_g$ can be computed from $v_{c} = Rv_g + T$.
$\partial \hat{n}_{v_i} / \partial v_g$ can be computed from Equations \ref{eq:vert_normal} and \ref{eq:face_normal}.

\section{SDF Regularization}
Two regularization terms are utilized during neural network training in order to encourage: (1) the inferred $\hat\phi$ values to resemble a true SDF and (2) smoothness (similar to \cite{li2023neuralangelo,rosu2023permutosdf}).
Notably, the smoothness regularizer behaves significantly better when $\hat{\phi}$ is closer to a true SDF.

\subsection{Eikonal Regularization}
Given a tetrahedron $t$ with vertices $u_k = (x_k, y_k, z_k)$ and inferred $\hat\phi_k$ values, $\hat{\phi}$ can be linearly approximated within the tetrahedron by writing
\begin{equation}\label{eq:sdf_linear_approx}
\hat{\phi}_k = ax_k + by_k + cz_k + d
\end{equation}
for each of the four vertices; then, the resulting $4\times 4$ linear system of equations can be solved to obtain the unknown coefficients $(a, b, c, d)$ leading to
\begin{equation}\label{eq:sdf_reg_phi}
|\nabla\hat{\phi}_t| = \sqrt{a^2 + b^2 + c^2}
\end{equation}
as the norm of the gradient.
Summing over tetrahedra leads to
\begin{equation}\label{eq:energy_1a}
E_{1a} = \frac 12 \sum_t (|\nabla\hat{\phi}_t| - 1)^2
\end{equation}
as the energy to be minimized.
The problem with Equation \ref{eq:energy_1a} (and similar approaches, such as \cite{gropp2020implicit,cai2022neural}) is that the chain rule moves the square root in Equation \ref{eq:sdf_reg_phi} to the denominator, potentially leading to NaNs/overflow; notably, even an exact SDF has $|\nabla\phi|=0$ at both extrema and pinching/merging saddles, and an inferred $\hat\phi$ can have $|\nabla\hat{\phi}|=0$ elsewhere as well.
This can be avoided by instead using
\begin{equation}
E_{1b} = \frac 12 \sum_t (|\nabla\hat{\phi}_t|^2 - 1)^2
\end{equation}
which still enforces $|\nabla\hat\phi_t| = 1$; alternatively,
\begin{equation}\label{eq:eikonal_loss}
E_{1c} = \frac 12 \sum_t \text{Volume}(t)(|\nabla\hat{\phi}_t|^2 - 1)^2
\end{equation}
scales the penalty on each tetrahedron by its volume.

\subsection{Motion by Mean Curvature}\label{sec:mean_curvature}
In order to encourage smoothness, we define an energy that when minimized results in motion by mean curvature. Following \cite{chan2001active,zhao1996variational}, the surface area can be calculated via
\begin{equation}
\int_{\Omega} |\nabla H(\phi(x,y,z))| dV
\end{equation}
where $H$ is a Heaviside function and $V$ is the volume; thus, on our tetrahedral mesh, we minimize 
\begin{equation}\label{eq:mmc}
E_{2} =  \sum_t |\nabla H(\hat{\phi})|\text{Volume}(t)
\end{equation}
using a smeared-out Heaviside Function
\begin{equation}
H(\hat{\phi}) = \begin{cases}
    0 &  \hat{\phi} < -\epsilon_H\\
    \frac 12 + \frac{\hat{\phi}}{2\epsilon_H} + \frac{1}{2\pi} \sin \left(\frac{\pi\hat{\phi} }{\epsilon_H}\right) & -\epsilon_H\leq\hat{\phi}\leq\epsilon_H\\
    1 & \hat{\phi} > \epsilon_H
\end{cases}
\end{equation}
where $\epsilon_H$, chosen as 1.5 times the average tetrahedral mesh edge length, determines the bandwidth of numerical smearing (see \cite{sussman1994level}).
$|\nabla H(\hat\phi)|$ is discretized by linearly approximating $H(\hat{\phi})$ in each tetrahedron along the lines of Equation \ref{eq:sdf_linear_approx} in order to obtain coefficients $(a,b,c,d)$ for use in the equivalent of Equation \ref{eq:sdf_reg_phi}.
In order to avoid division by small numbers, we ignore tetrahedra with $|\nabla H(\hat{\phi})|<10^{-8}$ in Equation \ref{eq:mmc} reasoning that $|\nabla H(\hat{\phi})|$ is small enough and thus $\hat\phi$ is smooth enough in such tetrahedra.

\section{Silhouette Losses}
Instead of striving to make the inverse rendering differentiable at silhouette boundaries (as in e.g.\ \cite{bangaru2022differentiable}), we introduce energies that force the silhouettes to match.

\subsection{Shrinking}
For pixels that overlap the inferred surface but not the ground truth surface, the interior of the inferred surface needs to shrink so that the corresponding triangles disappear.
For each tetrahedron mesh edge containing a vertex of a problematic triangle, the edge's parent tetrahedral mesh vertices are added to the set $U_{\text{shrink}}$ if they have negative SDF values; then, 
\begin{equation}\label{eq:shrink_loss}
    \mathcal{L}_{\text{shrink}} = \frac12 \sum_{k \in U_{\text{shrink}}} (\hat\phi_{k} -\epsilon_s)^2
\end{equation}
encourages those negative $\hat\phi_{k}$ values to target a positive $\epsilon_s=5\times10^{-3}$, which is chosen as half the average tetrahedral mesh edge length.

\subsection{Expanding}
For pixels that overlap the ground truth surface but not the inferred surface, the interior of the inferred surface needs to expand.
In order to determine where this expansion should occur, the implicit surface is temporarily inflated by changing the sign of the SDF at every tetrahedral mesh vertex with both $\hat\phi >0$ and a one-ring neighbor with $\hat\phi<0$ (e.g.\ by setting $\hat\phi_{temp} = -\epsilon_s$ at those vertices).
Next, the pixels that previously overlapped the ground truth surface but not the inferred surface and now overlap both the ground truth surface and the new inflated surface are identified.
For each tetrahedron mesh edge containing a vertex of a triangle corresponding to one of these pixels, the edge's parent tetrahedral mesh vertices are added to the set $U_{\text{expand}}$ if they had positive SDF values before inflation.
At this point, all of the temporary $\hat\phi_{temp}$ values are discarded and the original $\hat\phi$ values are restored.
Then, 
\begin{equation}\label{eq:expand_loss}
    \mathcal{L}_{\text{expand}} = \frac12 \sum_{k \in U_{\text{expand}}} (\hat\phi_{k}+\epsilon_s)^2
\end{equation}
encourages the positive $\hat\phi_{k}$ values to target $-\epsilon_s$.

\section{Experiments}
We first demonstrate (in Section \ref{sec:3d_gt}) that our network has the ability to reconstruct clothed humans when ground truth camera parameters and normal maps are known.
In Section \ref{sec:rgb_training}, we demonstrate that the network can be trained to reconstruct 3D geometry with increasing efficacy as the number of sparse views increases.
Subsequently (in Section \ref{sec:rgb_recon}), we extend this process to real-world RGB data (with no ground truth information) in order to demonstrate the ability to reconstruct 3D geometry using only network-inferred normal maps.
For the sake of comparison, we also present (in Section \ref{sec:comparisons}) the results we obtained using available implementations of other methods for single view and multiview reconstruction.

\subsection{Network Efficacy}\label{sec:3d_gt}
\begin{figure}[t]
    \centering
    \includegraphics[width=.8\linewidth]{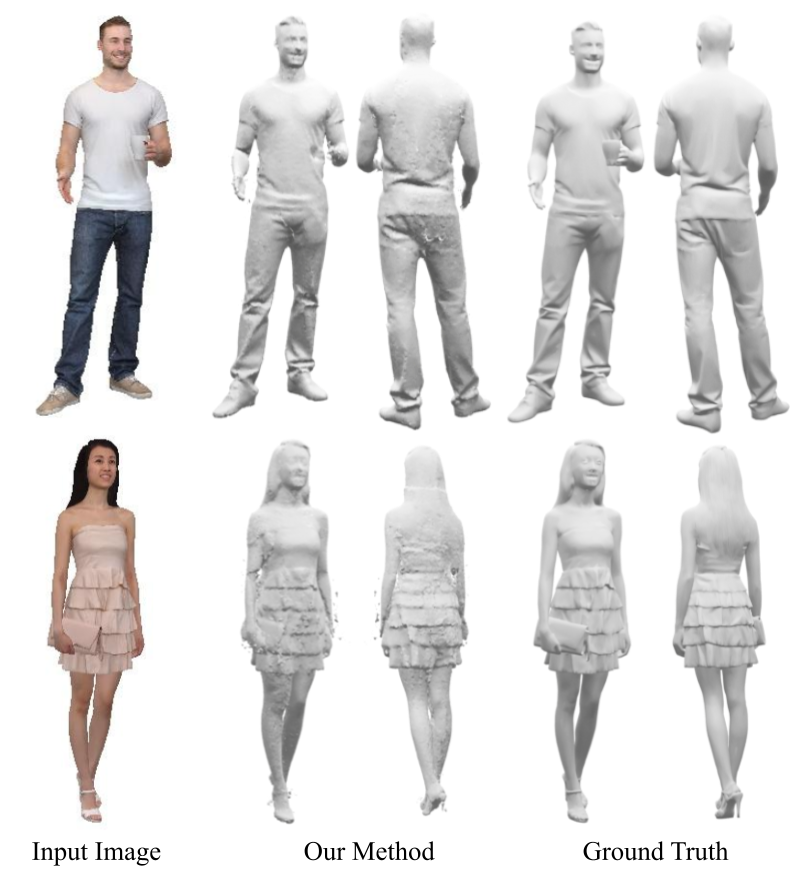}
    \caption{After training from 8 camera views, the input image in the first column results in the geometry shown in the second column. Note that the geometry is shown from novel views. For the sake of comparison, the ground truth  geometry is shown from the same novel views. See also Figure \ref{fig:pifuhd}.}
    \label{fig:rp_qual}
\end{figure}

Given ground truth 3D data from RenderPeople \cite{renderpeople}, we show that our network has the capacity and flexibility to reconstruct clothed humans from either a single image or multiple images.
Regardless of the number of input images, the network is trained by minimizing the normal map loss (Equation \ref{eq:normal_loss}), SDF regularization losses (Equations \ref{eq:eikonal_loss} and \ref{eq:mmc}), and silhouette losses (Equations \ref{eq:shrink_loss} and \ref{eq:expand_loss}).
In the multiview case, each image is considered individually (i.e.\ we treat multiview as a collection of single view examples).
Figure \ref{fig:rp_qual} shows an example of the results obtained by training our network on 8 camera views surrounding the person (as compared to the ground truth).

\begin{figure}[ht]
    \centering
    \includegraphics[width=0.9\linewidth]{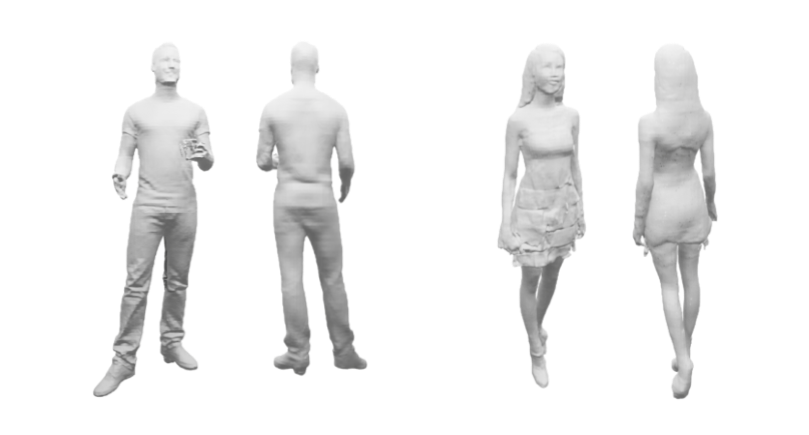}
    \caption{PIFuHD results, inferred using the input image in Figure \ref{fig:rp_qual} and shown from the same novel views (as in Figure \ref{fig:rp_qual}). We stress that these results were obtained using inference from a single image, and so one would not expect the same efficacy (especially from novel views); however, these images do help to calibrate what one might expect from state-of-the-art inference. The conclusion is that our network has the ability to output high-quality reconstructed geometry.}
    \label{fig:pifuhd}
\end{figure}

\subsection{Geometry Reconstruction}\label{sec:rgb_training}
To quantitatively evaluate the accuracy of our inferred results, we define a normal map error as
\begin{equation}
e_{normal} = \frac{1}{W\times H} \sum_p \left(\frac12 \left(1-\hat{n}_p \cdot n_p\right)\right)^2
\end{equation}
where the ground truth and predicted normals at pixel $p$ are $n_p$ and $\hat{n}_p$, respectively, and $\hat{n}_p \cdot n_p \in [-1,1]$ is replaced with $-1$ for pixels where the predicted and ground truth silhouettes do not overlap.
Note that normal maps do not uniquely determine scale/depth; thus, the reconstructed objects could erroneously move closer/further from the camera becoming smaller/larger in scale (while also undergoing distortion, since this scale variance is not self-similar).
In order to monitor this, we define
a depth map error as
\begin{equation}
e_{depth}  = \frac{1}{W\times H} \sum_p \left(\hat{d}_p - d_p\right)^2
\end{equation}
where $(\hat{d}_p - d_p)$ is replaced with the thickness of the tetrahedral mesh (0.2 meters) for pixels where the predicted and ground truth silhouettes do not overlap.

\begin{figure}[H]
\begin{subfigure}[b]{.32\linewidth}
\captionsetup{labelformat=empty}
    \includegraphics[width=\linewidth]{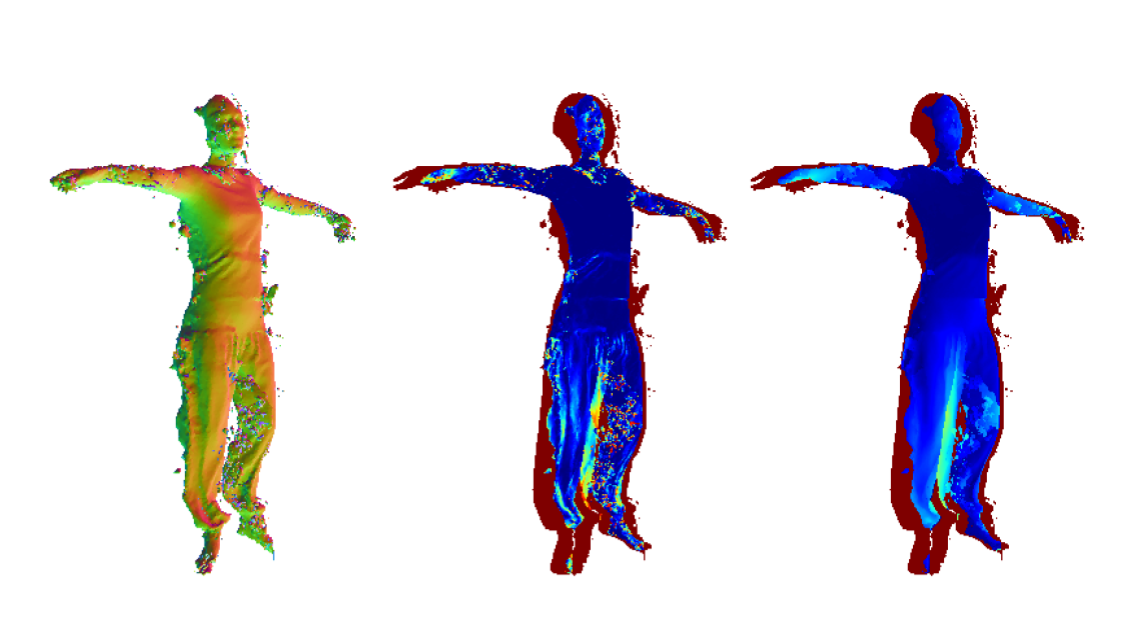}
    \caption{1 View}
    \label{fig:novel_view_a}
\end{subfigure}
\begin{subfigure}[b]{.32\linewidth}
\captionsetup{labelformat=empty}
    \includegraphics[width=\linewidth]{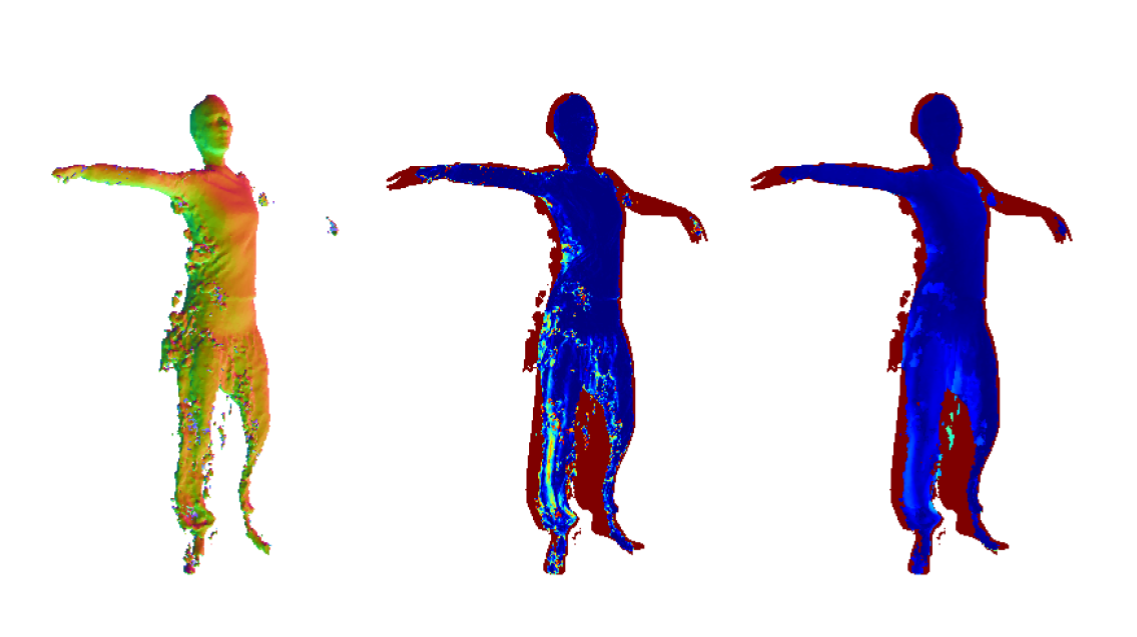}
    \caption{3 Views}
    \label{fig:novel_view_b}
\end{subfigure}
\begin{subfigure}[b]{.32\linewidth}
\captionsetup{labelformat=empty}
    \includegraphics[width=\linewidth]{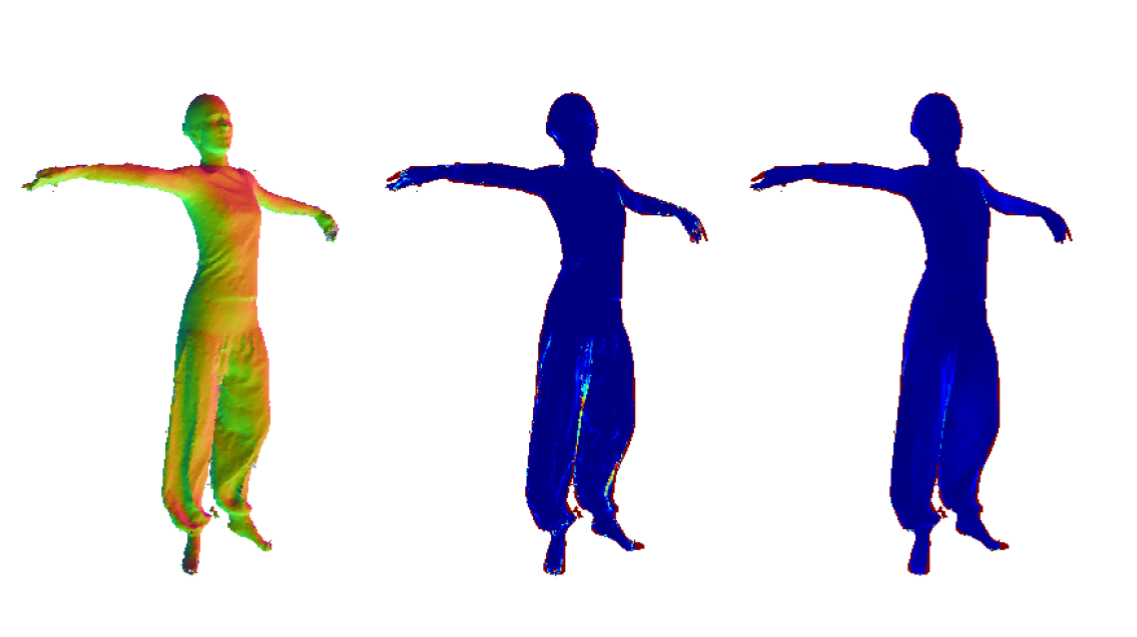}
    \caption{5 Views}
    \label{fig:novel_view_c}
\end{subfigure}
\caption{Models trained on an increasing number of camera views (inference from a novel view is shown). Each set of images show: (left) predicted triangle mesh and its normal map, (middle) normal errors (blue is zero, red is max), and (right) depth errors. See also Figure \ref{fig:3d_recon}.}
\label{fig:novel_view}
\end{figure}

Given ground truth 3D data from RenderPeople \cite{renderpeople}, we show how our network reconstructs 3D geometry with increasing efficacy as the number of sparse views increases.
Figure \ref{fig:novel_view} shows the inferred 3D geometry from a novel view, and Table \ref{tab:view_generalization} shows how per-pixel normal and depth errors decrease as the number of training views increases.
When the network is trained on only one view, there are no constraints on the side/back of the person; hence, the predicted geometry has a high degree of noise when rendered from novel views.
When trained with 5 views, the ground truth geometry is recovered with high accuracy.

\begin{table}[H]
\small
    \centering
    \begin{tabular}{rrrr}
    \toprule
     \# Views & Normals Error & Depth Error (m) \\
     \midrule
     1 & 0.0317 & 0.0012 \\
     3 & 0.0299 & 0.0011 \\
     5 & 0.0060 & 0.0002 \\
     \bottomrule
    \end{tabular}
    \caption{Quantitative metrics for three unseen test views spaced between the training views.}
    \label{tab:view_generalization}
\end{table}

\begin{figure}[ht]
\centering
\includegraphics[width=.49\linewidth]{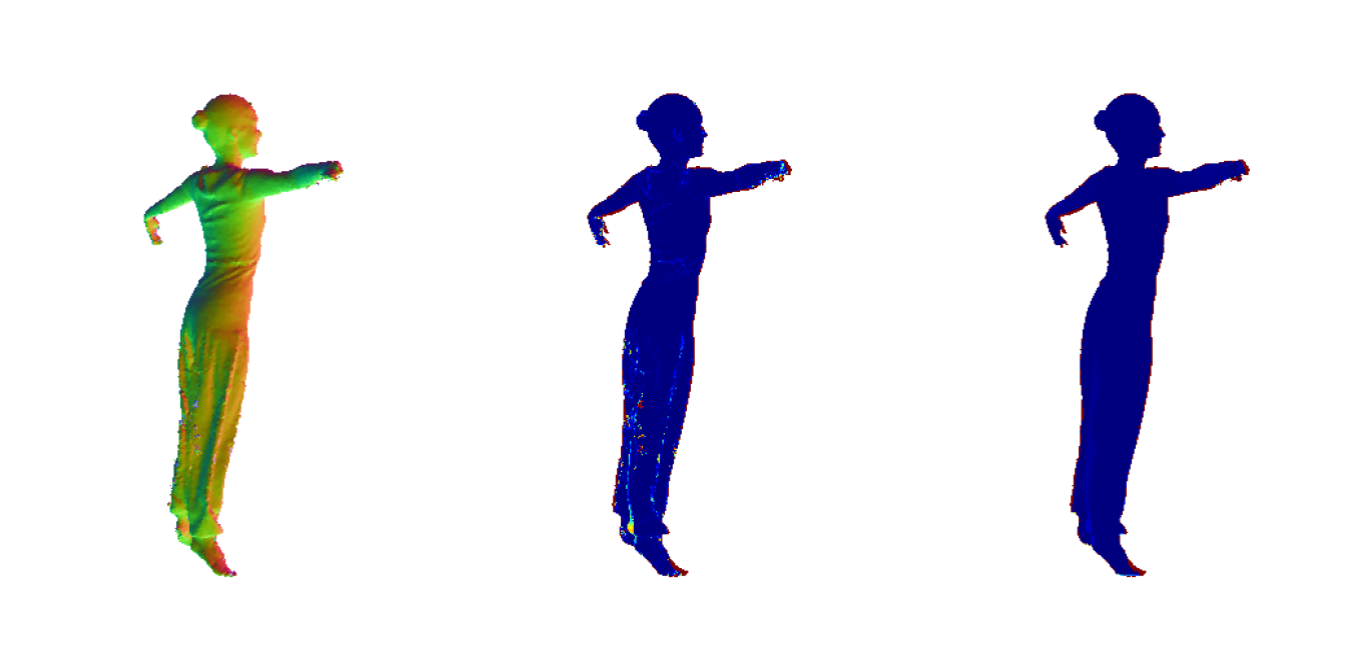}
\includegraphics[width=.49\linewidth]{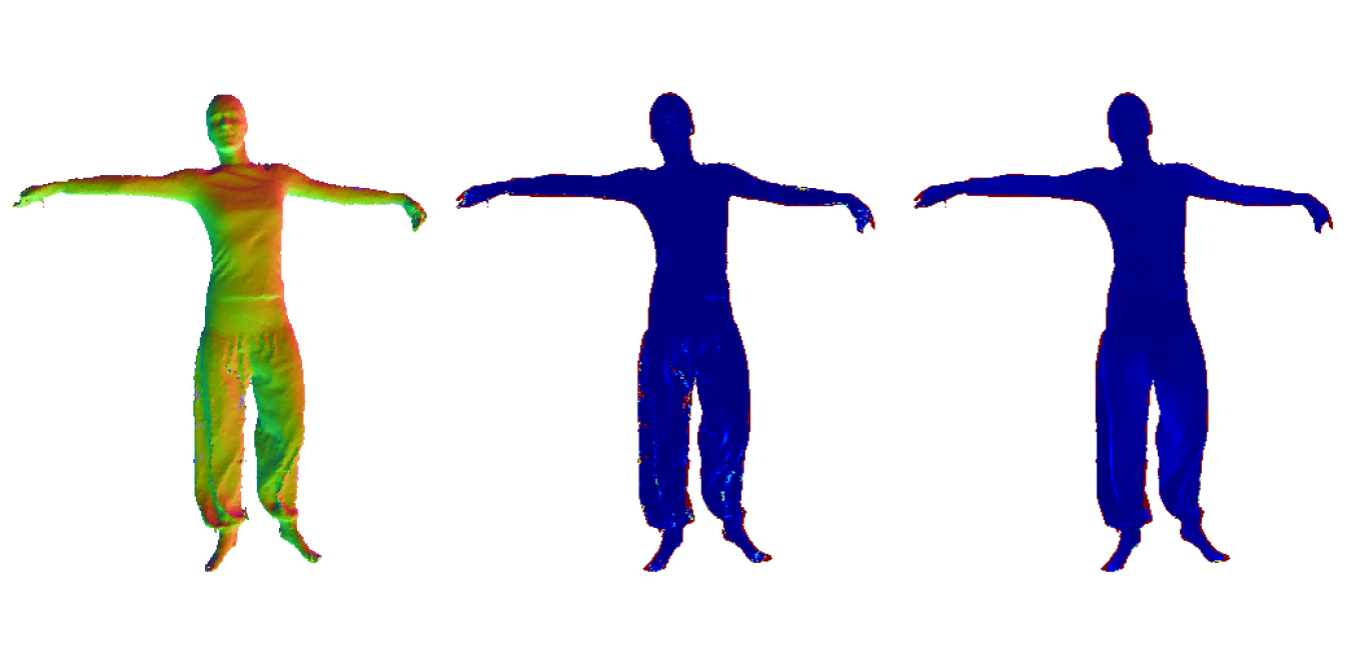}
\includegraphics[width=.49\linewidth]{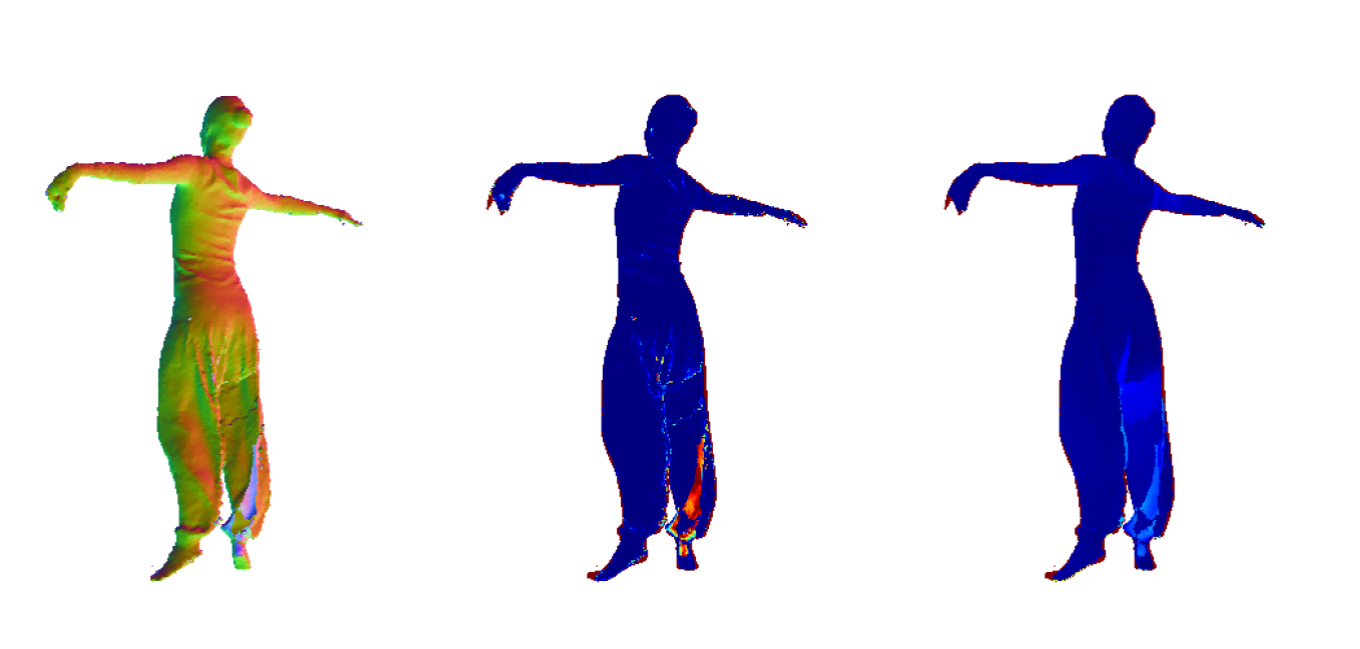}
\includegraphics[width=.49\linewidth]{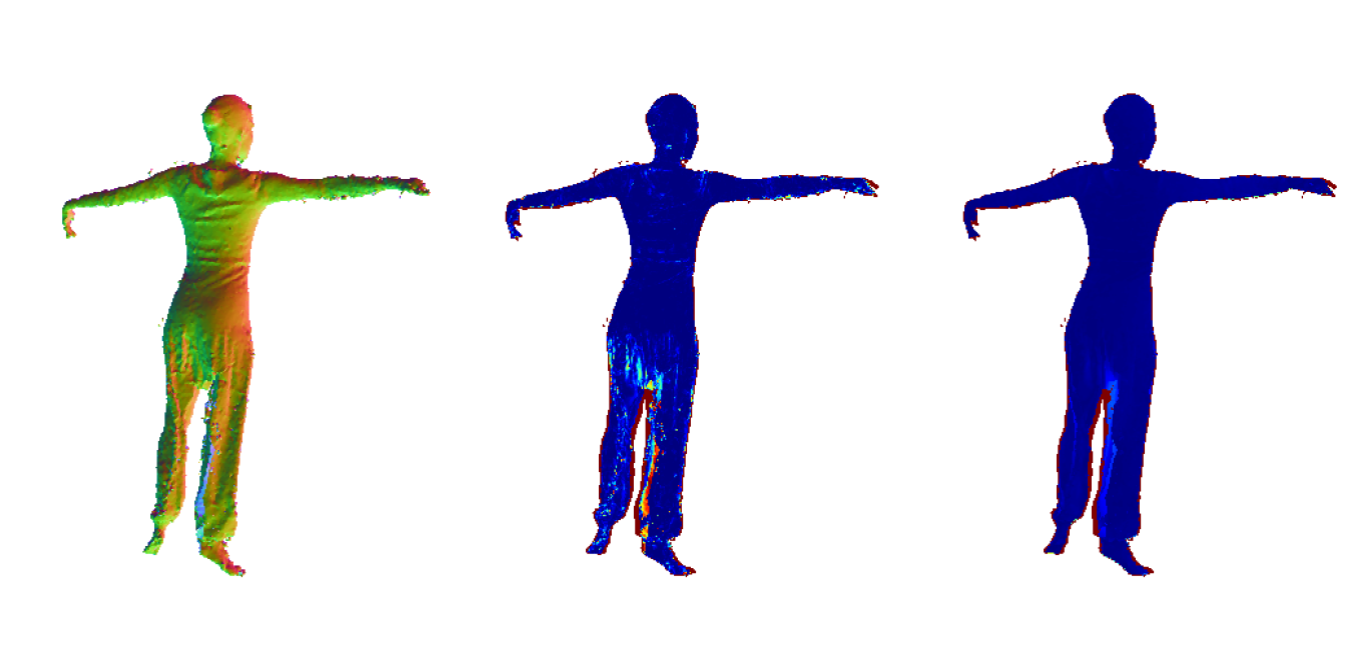}
\caption{Using the 5-view network from Figure \ref{fig:novel_view}, we illustrate inference for four more novel views.}
\label{fig:3d_recon}
\end{figure}

\subsection{Geometry Reconstruction from RGB Images}\label{sec:rgb_recon}
Here, we illustrate that our network can be used to reconstruct 3D geometry from monocular uncalibrated RGB images, without requiring any pretraining on scanned data (or any other informed initialization of the network parameters).
However, we do utilize a pretrained pix2pix network \cite{Wang_2018_CVPR}  (introduced in PIFuHD \cite{saito2020pifuhd}) to infer ground truth normal maps and note that pix2pix was trained on 3D ground truth geometry.
We do not consider this a severe limitation both because normal maps are easier to infer than 3D geometry and because there are other ways to obtain normal maps.

\begin{figure}[t]
\centering
\includegraphics[width=\linewidth]{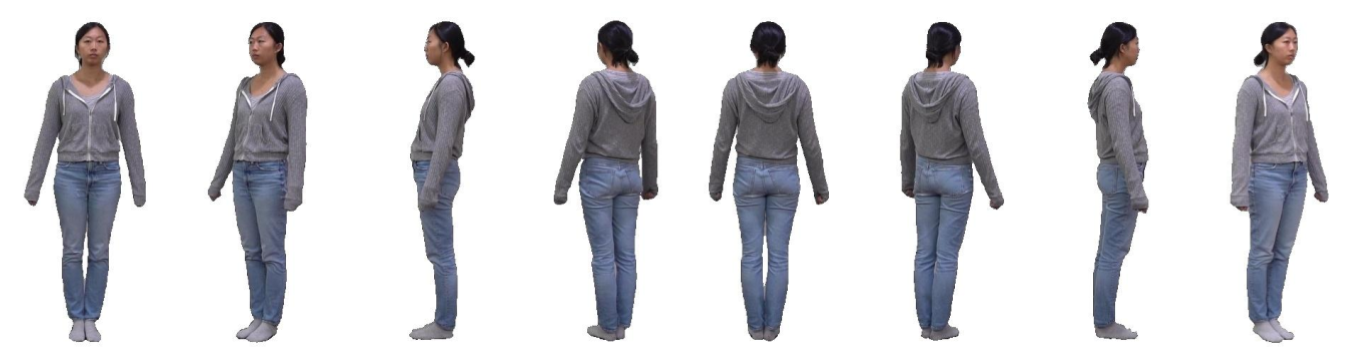}
\includegraphics[width=\linewidth]{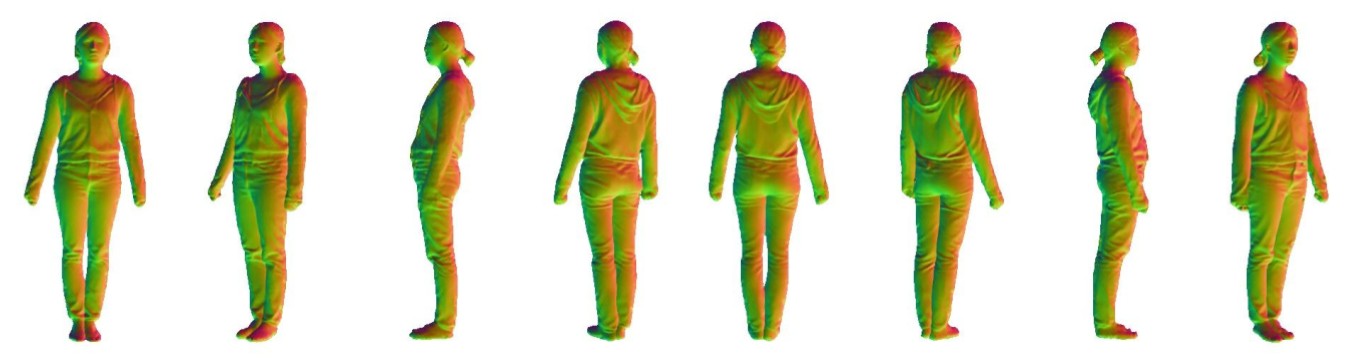}
\caption{RGB images and corresponding normal maps. Odd views were used for training (starting with first column), and even views are used for testing.}
\label{fig:rgb_results}
\end{figure}

\begin{figure}[t]
\centering
\includegraphics[width=\linewidth]{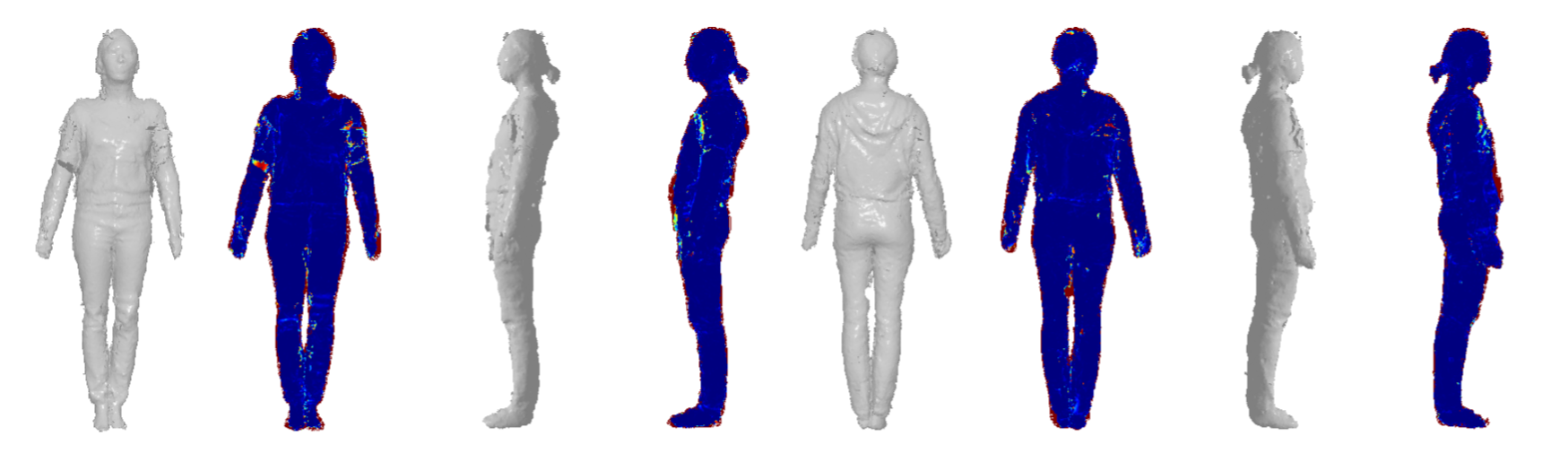}
\caption{Final per-view meshes alongside their corresponding normal map errors (blue is zero, red is max).}
\label{fig:rgb_per_view}
\end{figure}

\begin{figure}[t]
\centering
\includegraphics[width=\linewidth]{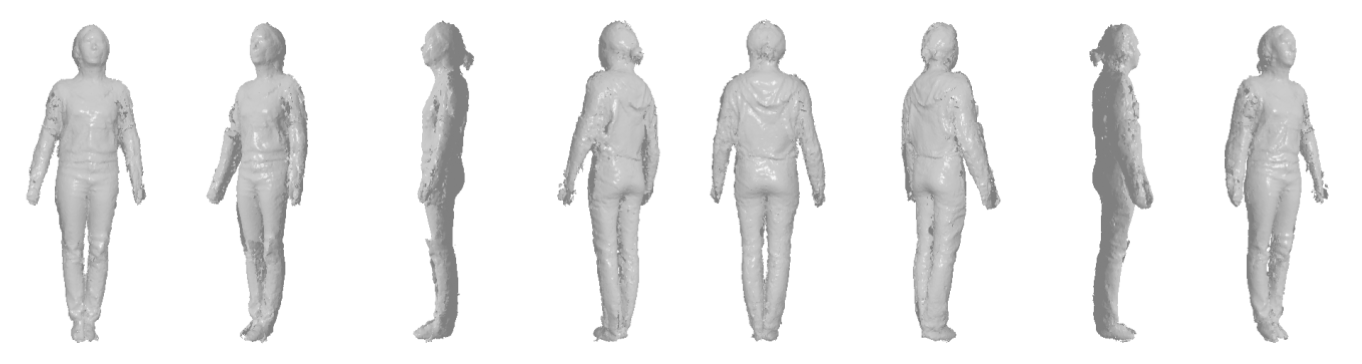}
\caption{Final triangle mesh shown from training (odd) and novel (even) views.}
\label{fig:rgb_final}
\end{figure}

First, we captured monocular video footage of a person in a static pose; then, a sparse number of frames were extracted and preprocessed by removing the background using \cite{carion2020end} and cropping to a square image.
The resulting images were then passed into pix2pix to obtain ``ground truth'' normal maps.
See Figure \ref{fig:rgb_results}.
Since estimated camera parameters will be prone to error, we refine a rough initialization iteratively.
At each iteration, we train the network and use Marching Tetrahedra to create a mesh inferenced off of the image for (and overfit to) each view; then, we use ICP \cite{besl1992method} to rigidly align all the meshes.
Although one could delete all the triangles and remesh the point cloud, we obtained better results by updating each camera to match the ICP rigid transform of its corresponding mesh.
The updated camera positions are then used to iteratively repeat the entire process.
Once the camera parameters converge, the network can be trained with an additional loss that encourages 3D consistency.
For a given camera view $c_0$, this loss is defined as
\begin{equation}\label{eq:multiview_loss_2}
    \mathcal{L}(\hat{\phi}_k) =  \sum_{c\neq c_0} \left\lVert \hat{\phi}_k - \hat{\phi}_k(c)\right\rVert
\end{equation}
where $\hat{\phi}_k(c)$ refers to the inferred SDF values obtained from using view $c$'s image.

\begin{figure*}[ht]
    \centering
\includegraphics[width=\linewidth]{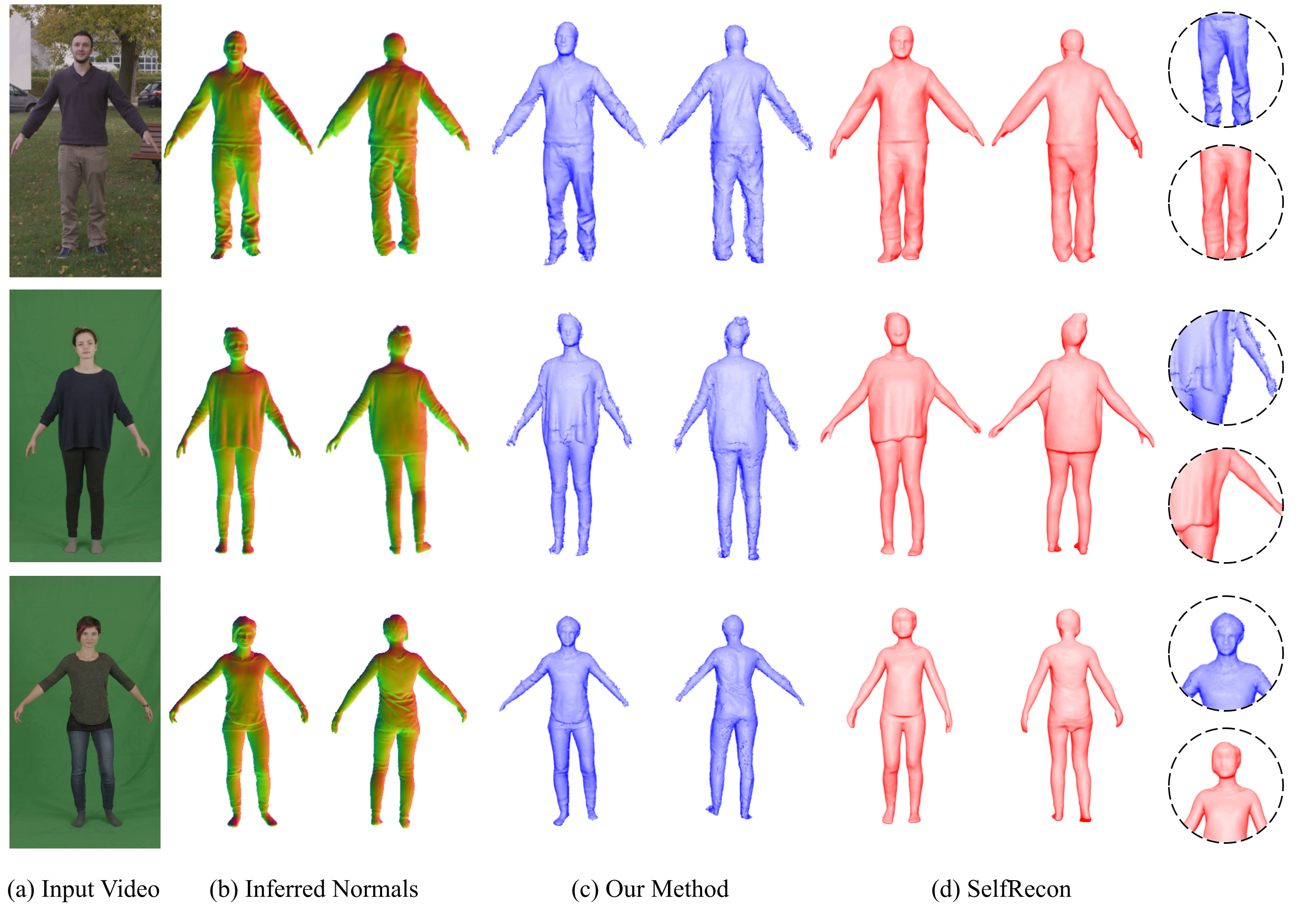}
    \caption{Predicted 3D geometry using our method (c) for videos from the PeopleSnapshot dataset (one frame is shown in (a)). The results from SelfRecon \cite{jiang2022selfrecon} are shown in (d). Note that the geometry is shown from novel views.}
    \label{fig:selfrecon}
\end{figure*}

The network obtained from the aforementioned process (to improve camera extrinsics) will tend to be less detailed on the back side of the mesh, since only the front side can be seen in any given input image; thus, after improving camera extrinsics, we proceed as follows.
Each view is fine-tuned with a regularizer that aims to keep $\phi$ close to that which was obtained using Equation \ref{eq:multiview_loss_2}; then, we delete any visible triangles that are not consistent with the normal map (within some tolerance).
See Figure \ref{fig:rgb_per_view}.
Since these are (actual) triangle meshes, it is trivial to load them into a suitable computer graphics application and align/resize the meshes in order to combine them into a single unified mesh. See Figure \ref{fig:rgb_final}.

\subsection{Comparisons}\label{sec:comparisons}
We quantitatively compare our reconstruction method to existing single view \cite{saito2020pifuhd,xiu2023econ} and multiview \cite{alldieck2018video,jiang2022selfrecon} reconstruction approaches using monocular videos from the People Snapshot Dataset \cite{alldieck2018video}.
Each video was captured with a fixed camera, and the subjects were asked to rotate while holding an A-pose.
We trained our network on four frames per video (front, back, and two side views) and subsequently deleted any visible triangles that are not consistent with the normal maps (within some tolerance, as in Section \ref{sec:rgb_recon}).
SelfRecon \cite{jiang2022selfrecon} and VideoAvatar \cite{alldieck2018video} were trained on all video frames.
For the single view approaches \cite{saito2020pifuhd,xiu2023econ}, we took the mesh predicted using the front or back-facing frame (whichever is closer to the test view) and scaled/rigidly aligned it using ICP to fit the corresponding SelfRecon mesh.
Table \ref{tab:quant_comparisons} compares the results obtained with each method to the PIFuHD inferred normal map.
SelfRecon has slightly more error and lacks detail compared to our approach (particularly around the face and wrinkles in the clothing).
See Figure \ref{fig:selfrecon}.
Notably, the runtime of our approach on a single NVIDIA 3090 GPU is at least 50$\times$ faster than SelfRecon, which takes over a day of training (per video) to achieve their published results (our network is trained for about 20 minutes).

\begin{table}[ht]
\small
    \centering
    \begin{tabular}{rrr}
    \toprule
     Method &  Average Normals Error &  STD\\
     \midrule
     ECON \cite{xiu2023econ} & 0.1918 & 0.376 \\
     PIFuHD \cite{saito2020pifuhd} & 0.1680 & 0.358\\
     VideoAvatar \cite{alldieck2018video} & 0.1342  & $0.325$ \\
     SelfRecon \cite{jiang2022selfrecon} & 0.0213 & $0.137$ \\
     \textbf{Our Method} & \textbf{0.0207} &  \textbf{0.111} \\
     \bottomrule
    \end{tabular}
    \caption{Normal map errors (computed over predicted foreground pixels) corresponding to four test views, averaged over the exam- ples shown in Figure \ref{fig:selfrecon}.}
    \label{tab:quant_comparisons}
\end{table}
\label{sec:experiments}

\section{Conclusion}

Although image-based reconstruction can be solved as an inverse problem, regularization is required in order to address issues with noise.
Parameterized models (such as SMPL \cite{loper2015smpl} or 3DMM \cite{egger20203d}) provide for such regularization.
We choose a neural network to parameterize our reconstruction where regularization is provided by having a limited number of network parameters.
Our network aims to convert images from any view direction into a unique implicit surface, regardless of the view direction (similar in spirit to how the human brain process visual input); in fact, our eyes discern relative distance (similar to normal maps) more proficiently than they discern raw distance.


In summary, we present a weakly-supervised method for clothed human reconstruction by leveraging 2D normal maps as the supervisory signal during neural network training.
In order to train a learned model that can infer high-frequency cloth and body geometry without any ground truth 3D data, our proposed approach builds on strong geometric priors for modeling and rendering.
Our results reinforce the notion that less training data is required to train networks that infer normal maps than to train networks that infer 3D geometry (in agreement with ECON \cite{xiu2023econ}). 
This means that working to improve the efficacy of network-inferred normal maps (and using the results for 3D reconstruction, as in Section \ref{sec:rgb_recon}) is likely to be more productive than working to obtain (via expensive 3D scanning) the excessive amount of ground truth data required to train a network to inference 3D geometry directly.
Moreover, the process outlined in Section \ref{sec:rgb_recon} provides an alternative mechanism (significantly cheaper than 3D scanning) for acquiring the ground truth data required to train a network to inference 3D geometry directly.

\section*{Acknowledgements}
Research supported in part by ONR N00014-19-1-2285, ONR N00014-21-1-2771. We
would like to thank Reza and Behzad at ONR for supporting our efforts into machine learning.
This work was also supported by JSPS KAKENHI Grant Number JP23H03439.
J. W. was supported in part by the Gerald J. Lieberman Graduate Fellowship, the NSF Mathematical Sciences Postdoctoral Fellowship, and the UC President's Postdoctoral Fellowship.
\label{sec:conclusion}


{
\small
\bibliographystyle{ieeenat_fullname}
\bibliography{mainbib}

\begin{thebibliography}{104}
\providecommand{\natexlab}[1]{#1}
\providecommand{\url}[1]{\texttt{#1}}
\expandafter\ifx\csname urlstyle\endcsname\relax
  \providecommand{\doi}[1]{doi: #1}\else
  \providecommand{\doi}{doi: \begingroup \urlstyle{rm}\Url}\fi

\bibitem[Alldieck et~al.(2018)Alldieck, Magnor, Xu, Theobalt, and
  Pons-Moll]{alldieck2018video}
Thiemo Alldieck, Marcus Magnor, Weipeng Xu, Christian Theobalt, and Gerard
  Pons-Moll.
\newblock Video based reconstruction of 3d people models.
\newblock In \emph{Proceedings of the IEEE Conference on Computer Vision and
  Pattern Recognition}, pages 8387--8397, 2018.

\bibitem[Andriluka et~al.(2014)Andriluka, Pishchulin, Gehler, and
  Schiele]{andriluka20142d}
Mykhaylo Andriluka, Leonid Pishchulin, Peter Gehler, and Bernt Schiele.
\newblock 2d human pose estimation: New benchmark and state of the art
  analysis.
\newblock In \emph{Proceedings of the IEEE Conference on computer Vision and
  Pattern Recognition}, pages 3686--3693, 2014.

\bibitem[Badki et~al.(2020)Badki, Troccoli, Kim, Kautz, Sen, and
  Gallo]{badki2020bi3d}
Abhishek Badki, Alejandro Troccoli, Kihwan Kim, Jan Kautz, Pradeep Sen, and
  Orazio Gallo.
\newblock Bi3d: Stereo depth estimation via binary classifications.
\newblock In \emph{Proceedings of the IEEE/CVF Conference on Computer Vision
  and Pattern Recognition}, pages 1600--1608, 2020.

\bibitem[Bangaru et~al.(2022)Bangaru, Gharbi, Luan, Li, Sunkavalli, Hasan, Bi,
  Xu, Bernstein, and Durand]{bangaru2022differentiable}
Sai~Praveen Bangaru, Micha{\"e}l Gharbi, Fujun Luan, Tzu-Mao Li, Kalyan
  Sunkavalli, Milos Hasan, Sai Bi, Zexiang Xu, Gilbert Bernstein, and Fredo
  Durand.
\newblock Differentiable rendering of neural sdfs through reparameterization.
\newblock In \emph{SIGGRAPH Asia 2022 Conference Papers}, pages 1--9, 2022.

\bibitem[Besl and McKay(1992)]{besl1992method}
Paul~J Besl and Neil~D McKay.
\newblock Method for registration of 3-d shapes.
\newblock In \emph{Sensor fusion IV: control paradigms and data structures},
  pages 586--606. Spie, 1992.

\bibitem[Cai et~al.(2022)Cai, Feng, Feng, Wang, and Zhang]{cai2022neural}
Hongrui Cai, Wanquan Feng, Xuetao Feng, Yan Wang, and Juyong Zhang.
\newblock Neural surface reconstruction of dynamic scenes with monocular rgb-d
  camera.
\newblock \emph{arXiv preprint arXiv:2206.15258}, 2022.

\bibitem[Cao et~al.(2023)Cao, Han, and Wong]{cao2023sesdf}
Yukang Cao, Kai Han, and Kwan-Yee~K Wong.
\newblock Sesdf: Self-evolved signed distance field for implicit 3d clothed
  human reconstruction.
\newblock In \emph{Proceedings of the IEEE/CVF Conference on Computer Vision
  and Pattern Recognition}, pages 4647--4657, 2023.

\bibitem[Carion et~al.(2020)Carion, Massa, Synnaeve, Usunier, Kirillov, and
  Zagoruyko]{carion2020end}
Nicolas Carion, Francisco Massa, Gabriel Synnaeve, Nicolas Usunier, Alexander
  Kirillov, and Sergey Zagoruyko.
\newblock End-to-end object detection with transformers.
\newblock In \emph{European conference on computer vision}, pages 213--229.
  Springer, 2020.

\bibitem[Chan et~al.(2022)Chan, Lin, Zhao, and Lin]{chan2022integratedpifu}
Kennard~Yanting Chan, Guosheng Lin, Haiyu Zhao, and Weisi Lin.
\newblock Integratedpifu: Integrated pixel aligned implicit function for
  single-view human reconstruction.
\newblock In \emph{Computer Vision--ECCV 2022: 17th European Conference, Tel
  Aviv, Israel, October 23--27, 2022, Proceedings, Part II}, pages 328--344.
  Springer, 2022.

\bibitem[Chan and Vese(2001)]{chan2001active}
Tony~F Chan and Luminita~A Vese.
\newblock Active contours without edges.
\newblock \emph{IEEE Transactions on image processing}, 10\penalty0
  (2):\penalty0 266--277, 2001.

\bibitem[Chen et~al.(2022)Chen, Jiang, Song, Yang, Black, Geiger, and
  Hilliges]{chen2022gdna}
Xu Chen, Tianjian Jiang, Jie Song, Jinlong Yang, Michael~J Black, Andreas
  Geiger, and Otmar Hilliges.
\newblock gdna: Towards generative detailed neural avatars.
\newblock In \emph{Proceedings of the IEEE/CVF Conference on Computer Vision
  and Pattern Recognition}, pages 20427--20437, 2022.

\bibitem[Corona et~al.(2022)Corona, Zanfir, Alldieck, Bazavan, Zanfir, and
  Sminchisescu]{corona2022structured}
Enric Corona, Mihai Zanfir, Thiemo Alldieck, Eduard~Gabriel Bazavan, Andrei
  Zanfir, and Cristian Sminchisescu.
\newblock Structured 3d features for reconstructing relightable and animatable
  avatars.
\newblock \emph{arXiv preprint arXiv:2212.06820}, 2022.

\bibitem[De~Luigi et~al.(2023)De~Luigi, Li, Guillard, Salzmann, and
  Fua]{de2023drapenet}
Luca De~Luigi, Ren Li, Beno{\^\i}t Guillard, Mathieu Salzmann, and Pascal Fua.
\newblock Drapenet: Garment generation and self-supervised draping.
\newblock In \emph{Proceedings of the IEEE/CVF Conference on Computer Vision
  and Pattern Recognition}, pages 1451--1460, 2023.

\bibitem[Dong et~al.(2022{\natexlab{a}})Dong, Fang, Guo, Peng, Shuai, Zhou, and
  Bao]{dong2022totalselfscan}
Junting Dong, Qi Fang, Yudong Guo, Sida Peng, Qing Shuai, Xiaowei Zhou, and
  Hujun Bao.
\newblock Totalselfscan: Learning full-body avatars from self-portrait videos
  of faces, hands, and bodies.
\newblock \emph{Advances in Neural Information Processing Systems},
  35:\penalty0 13654--13667, 2022{\natexlab{a}}.

\bibitem[Dong et~al.(2022{\natexlab{b}})Dong, Xu, Duan, Bao, Xu, and
  Lau]{dong2022geometry}
Zheng Dong, Ke Xu, Ziheng Duan, Hujun Bao, Weiwei Xu, and Rynson Lau.
\newblock Geometry-aware two-scale pifu representation for human
  reconstruction.
\newblock \emph{Advances in Neural Information Processing Systems},
  35:\penalty0 31130--31144, 2022{\natexlab{b}}.

\bibitem[Egger et~al.(2020)Egger, Smith, Tewari, Wuhrer, Zollhoefer, Beeler,
  Bernard, Bolkart, Kortylewski, Romdhani, et~al.]{egger20203d}
Bernhard Egger, William~AP Smith, Ayush Tewari, Stefanie Wuhrer, Michael
  Zollhoefer, Thabo Beeler, Florian Bernard, Timo Bolkart, Adam Kortylewski,
  Sami Romdhani, et~al.
\newblock 3d morphable face models: past, present, and future.
\newblock \emph{ACM Transactions on Graphics (ToG)}, 39\penalty0 (5):\penalty0
  1--38, 2020.

\bibitem[Feng et~al.(2022{\natexlab{a}})Feng, Liu, Lai, Yang, and
  Li]{feng2022fof}
Qiao Feng, Yebin Liu, Yu-Kun Lai, Jingyu Yang, and Kun Li.
\newblock Fof: learning fourier occupancy field for monocular real-time human
  reconstruction.
\newblock \emph{Advances in Neural Information Processing Systems},
  35:\penalty0 7397--7409, 2022{\natexlab{a}}.

\bibitem[Feng et~al.(2022{\natexlab{b}})Feng, Yang, Pollefeys, Black, and
  Bolkart]{feng2022capturing}
Yao Feng, Jinlong Yang, Marc Pollefeys, Michael~J Black, and Timo Bolkart.
\newblock Capturing and animation of body and clothing from monocular video.
\newblock \emph{arXiv preprint arXiv:2210.01868}, 2022{\natexlab{b}}.

\bibitem[Gabeur et~al.(2019)Gabeur, Franco, Martin, Schmid, and
  Rogez]{gabeur2019moulding}
Valentin Gabeur, Jean-S{\'e}bastien Franco, Xavier Martin, Cordelia Schmid, and
  Gregory Rogez.
\newblock Moulding humans: Non-parametric 3d human shape estimation from single
  images.
\newblock In \emph{Proceedings of the IEEE/CVF International Conference on
  Computer Vision}, pages 2232--2241, 2019.

\bibitem[Godard et~al.(2019)Godard, Mac~Aodha, Firman, and
  Brostow]{godard2019digging}
Cl{\'e}ment Godard, Oisin Mac~Aodha, Michael Firman, and Gabriel~J Brostow.
\newblock Digging into self-supervised monocular depth estimation.
\newblock In \emph{Proceedings of the IEEE/CVF international conference on
  computer vision}, pages 3828--3838, 2019.

\bibitem[Goel et~al.(2023)Goel, Pavlakos, Rajasegaran, Kanazawa*, and
  Malik*]{goel2023humans}
Shubham Goel, Georgios Pavlakos, Jathushan Rajasegaran, Angjoo Kanazawa*, and
  Jitendra Malik*.
\newblock Humans in 4{D}: Reconstructing and tracking humans with transformers.
\newblock In \emph{International Conference on Computer Vision (ICCV)}, 2023.

\bibitem[Gropp et~al.(2020)Gropp, Yariv, Haim, Atzmon, and
  Lipman]{gropp2020implicit}
Amos Gropp, Lior Yariv, Niv Haim, Matan Atzmon, and Yaron Lipman.
\newblock Implicit geometric regularization for learning shapes.
\newblock \emph{arXiv preprint arXiv:2002.10099}, 2020.

\bibitem[Guo et~al.(2023)Guo, Jiang, Chen, Song, and
  Hilliges]{guo2023vid2avatar}
Chen Guo, Tianjian Jiang, Xu Chen, Jie Song, and Otmar Hilliges.
\newblock Vid2avatar: 3d avatar reconstruction from videos in the wild via
  self-supervised scene decomposition.
\newblock In \emph{Proceedings of the IEEE/CVF Conference on Computer Vision
  and Pattern Recognition}, pages 12858--12868, 2023.

\bibitem[Habermann et~al.(2019)Habermann, Xu, Zollhoefer, Pons-Moll, and
  Theobalt]{habermann2019livecap}
Marc Habermann, Weipeng Xu, Michael Zollhoefer, Gerard Pons-Moll, and Christian
  Theobalt.
\newblock Livecap: Real-time human performance capture from monocular video.
\newblock \emph{ACM Transactions on Graphics (TOG)}, 38\penalty0 (2):\penalty0
  14, 2019.

\bibitem[Habermann et~al.(2020)Habermann, Xu, Zollhofer, Pons-Moll, and
  Theobalt]{habermann2020deepcap}
Marc Habermann, Weipeng Xu, Michael Zollhofer, Gerard Pons-Moll, and Christian
  Theobalt.
\newblock Deepcap: Monocular human performance capture using weak supervision.
\newblock In \emph{Proceedings of the IEEE/CVF Conference on Computer Vision
  and Pattern Recognition}, pages 5052--5063, 2020.

\bibitem[Han et~al.(2023)Han, Park, Yoon, Kang, Park, and Jeon]{han2023high}
Sang-Hun Han, Min-Gyu Park, Ju~Hong Yoon, Ju-Mi Kang, Young-Jae Park, and
  Hae-Gon Jeon.
\newblock High-fidelity 3d human digitization from single 2k resolution images.
\newblock In \emph{Proceedings of the IEEE/CVF Conference on Computer Vision
  and Pattern Recognition}, pages 12869--12879, 2023.

\bibitem[Hong et~al.(2021)Hong, Zhang, Jiang, Guo, Liu, and
  Bao]{hong2021stereopifu}
Yang Hong, Juyong Zhang, Boyi Jiang, Yudong Guo, Ligang Liu, and Hujun Bao.
\newblock Stereopifu: Depth aware clothed human digitization via stereo vision.
\newblock In \emph{Proceedings of the IEEE/CVF Conference on Computer Vision
  and Pattern Recognition}, pages 535--545, 2021.

\bibitem[I{\c{s}}{\i}k et~al.(2023)I{\c{s}}{\i}k, R{\"u}nz, Georgopoulos,
  Khakhulin, Starck, Agapito, and Nie{\ss}ner]{icsik2023humanrf}
Mustafa I{\c{s}}{\i}k, Martin R{\"u}nz, Markos Georgopoulos, Taras Khakhulin,
  Jonathan Starck, Lourdes Agapito, and Matthias Nie{\ss}ner.
\newblock Humanrf: High-fidelity neural radiance fields for humans in motion.
\newblock \emph{arXiv preprint arXiv:2305.06356}, 2023.

\bibitem[Jiang et~al.(2022{\natexlab{a}})Jiang, Hong, Bao, and
  Zhang]{jiang2022selfrecon}
Boyi Jiang, Yang Hong, Hujun Bao, and Juyong Zhang.
\newblock Selfrecon: Self reconstruction your digital avatar from monocular
  video.
\newblock In \emph{Proceedings of the IEEE/CVF Conference on Computer Vision
  and Pattern Recognition}, pages 5605--5615, 2022{\natexlab{a}}.

\bibitem[Jiang et~al.(2023)Jiang, Chen, Song, and
  Hilliges]{jiang2023instantavatar}
Tianjian Jiang, Xu Chen, Jie Song, and Otmar Hilliges.
\newblock Instantavatar: Learning avatars from monocular video in 60 seconds.
\newblock In \emph{Proceedings of the IEEE/CVF Conference on Computer Vision
  and Pattern Recognition}, pages 16922--16932, 2023.

\bibitem[Jiang et~al.(2022{\natexlab{b}})Jiang, Yi, Samei, Tuzel, and
  Ranjan]{jiang2022neuman}
Wei Jiang, Kwang~Moo Yi, Golnoosh Samei, Oncel Tuzel, and Anurag Ranjan.
\newblock Neuman: Neural human radiance field from a single video.
\newblock In \emph{Computer Vision--ECCV 2022: 17th European Conference, Tel
  Aviv, Israel, October 23--27, 2022, Proceedings, Part XXXII}, pages 402--418.
  Springer, 2022{\natexlab{b}}.

\bibitem[Jinka et~al.(2023)Jinka, Srivastava, Pokhariya, Sharma, and
  Narayanan]{jinka2023sharp}
Sai~Sagar Jinka, Astitva Srivastava, Chandradeep Pokhariya, Avinash Sharma, and
  PJ Narayanan.
\newblock Sharp: Shape-aware reconstruction of people in loose clothing.
\newblock \emph{International Journal of Computer Vision}, 131\penalty0
  (4):\penalty0 918--937, 2023.

\bibitem[Jun et~al.(2021)Jun, Lee, Lee, and Kim]{jun2021monocular}
Jinyoung Jun, Jae-Han Lee, Chul Lee, and Chang-Su Kim.
\newblock Monocular human depth estimation via pose estimation.
\newblock \emph{IEEE Access}, 9:\penalty0 151444--151457, 2021.

\bibitem[Kanazawa et~al.(2018)Kanazawa, Black, Jacobs, and
  Malik]{kanazawa2018end}
Angjoo Kanazawa, Michael~J Black, David~W Jacobs, and Jitendra Malik.
\newblock End-to-end recovery of human shape and pose.
\newblock In \emph{Proceedings of the IEEE conference on computer vision and
  pattern recognition}, pages 7122--7131, 2018.

\bibitem[Kim et~al.(2022)Kim, Nam, Kim, Park, and Lee]{kim2022laplacianfusion}
Hyomin Kim, Hyeonseo Nam, Jungeon Kim, Jaesik Park, and Seungyong Lee.
\newblock Laplacianfusion: Detailed 3d clothed-human body reconstruction.
\newblock \emph{ACM Transactions on Graphics (TOG)}, 41\penalty0 (6):\penalty0
  1--14, 2022.

\bibitem[Kim et~al.(2023)Kim, Gwon, Park, Kwon, Um, and Kim]{kim2023sampling}
Jeonghwan Kim, Mi-Gyeong Gwon, Hyunwoo Park, Hyukmin Kwon, Gi-Mun Um, and
  Wonjun Kim.
\newblock Sampling is matter: Point-guided 3d human mesh reconstruction.
\newblock In \emph{Proceedings of the IEEE/CVF Conference on Computer Vision
  and Pattern Recognition}, pages 12880--12889, 2023.

\bibitem[Kocabas et~al.(2020)Kocabas, Athanasiou, and Black]{kocabas2020vibe}
Muhammed Kocabas, Nikos Athanasiou, and Michael~J Black.
\newblock Vibe: Video inference for human body pose and shape estimation.
\newblock In \emph{Proceedings of the IEEE/CVF Conference on Computer Vision
  and Pattern Recognition}, pages 5253--5263, 2020.

\bibitem[Kolotouros et~al.(2021)Kolotouros, Pavlakos, Jayaraman, and
  Daniilidis]{kolotouros2021probabilistic}
Nikos Kolotouros, Georgios Pavlakos, Dinesh Jayaraman, and Kostas Daniilidis.
\newblock Probabilistic modeling for human mesh recovery.
\newblock In \emph{Proceedings of the IEEE/CVF International Conference on
  Computer Vision}, pages 11605--11614, 2021.

\bibitem[Kusupati et~al.(2020)Kusupati, Cheng, Chen, and
  Su]{kusupati2020normal}
Uday Kusupati, Shuo Cheng, Rui Chen, and Hao Su.
\newblock Normal assisted stereo depth estimation.
\newblock In \emph{Proceedings of the IEEE/CVF Conference on Computer Vision
  and Pattern Recognition}, pages 2189--2199, 2020.

\bibitem[Lee et~al.(2018)Lee, Hyde, Bao, and Fedkiw]{lee2018skinned}
Minjae Lee, David Hyde, Michael Bao, and Ronald Fedkiw.
\newblock A skinned tetrahedral mesh for hair animation and hair-water
  interaction.
\newblock \emph{IEEE transactions on visualization and computer graphics},
  25\penalty0 (3):\penalty0 1449--1459, 2018.

\bibitem[Lee et~al.(2019)Lee, Hyde, Li, and Fedkiw]{lee2019robust}
Minjae Lee, David Hyde, Kevin Li, and Ronald Fedkiw.
\newblock A robust volume conserving method for character-water interaction.
\newblock In \emph{Proceedings of the 18th annual ACM SIGGRAPH/Eurographics
  Symposium on Computer Animation}, pages 1--12, 2019.

\bibitem[Li et~al.(2022)Li, Zheng, Zhang, Ji, and Liu]{li2022avatarcap}
Zhe Li, Zerong Zheng, Hongwen Zhang, Chaonan Ji, and Yebin Liu.
\newblock Avatarcap: Animatable avatar conditioned monocular human volumetric
  capture.
\newblock In \emph{Computer Vision--ECCV 2022: 17th European Conference, Tel
  Aviv, Israel, October 23--27, 2022, Proceedings, Part I}, pages 322--341.
  Springer, 2022.

\bibitem[Li et~al.(2023)Li, M{\"u}ller, Evans, Taylor, Unberath, Liu, and
  Lin]{li2023neuralangelo}
Zhaoshuo Li, Thomas M{\"u}ller, Alex Evans, Russell~H Taylor, Mathias Unberath,
  Ming-Yu Liu, and Chen-Hsuan Lin.
\newblock Neuralangelo: High-fidelity neural surface reconstruction.
\newblock In \emph{Proceedings of the IEEE/CVF Conference on Computer Vision
  and Pattern Recognition}, pages 8456--8465, 2023.

\bibitem[Liao et~al.(2018)Liao, Donne, and Geiger]{liao2018deep}
Yiyi Liao, Simon Donne, and Andreas Geiger.
\newblock Deep marching cubes: Learning explicit surface representations.
\newblock In \emph{Proceedings of the IEEE Conference on Computer Vision and
  Pattern Recognition}, pages 2916--2925, 2018.

\bibitem[Lin et~al.(2022)Lin, Zhang, Zheng, Shao, and Liu]{lin2022learning}
Siyou Lin, Hongwen Zhang, Zerong Zheng, Ruizhi Shao, and Yebin Liu.
\newblock Learning implicit templates for point-based clothed human modeling.
\newblock In \emph{Computer Vision--ECCV 2022: 17th European Conference, Tel
  Aviv, Israel, October 23--27, 2022, Proceedings, Part III}, pages 210--228.
  Springer, 2022.

\bibitem[Liu et~al.(2022)Liu, Bao, Sun, and Mei]{liu2022recent}
Wu Liu, Qian Bao, Yu Sun, and Tao Mei.
\newblock Recent advances of monocular 2d and 3d human pose estimation: a deep
  learning perspective.
\newblock \emph{ACM Computing Surveys}, 55\penalty0 (4):\penalty0 1--41, 2022.

\bibitem[Loper et~al.(2015)Loper, Mahmood, Romero, Pons-Moll, and
  Black]{loper2015smpl}
Matthew Loper, Naureen Mahmood, Javier Romero, Gerard Pons-Moll, and Michael~J
  Black.
\newblock Smpl: A skinned multi-person linear model.
\newblock \emph{ACM transactions on graphics (TOG)}, 34\penalty0 (6):\penalty0
  1--16, 2015.

\bibitem[Lorensen and Cline(1987)]{lorensen1987marching}
William~E Lorensen and Harvey~E Cline.
\newblock Marching cubes: A high resolution 3d surface construction algorithm.
\newblock \emph{ACM siggraph computer graphics}, 21\penalty0 (4):\penalty0
  163--169, 1987.

\bibitem[Ma et~al.(2021)Ma, Yang, Tang, and Black]{ma2021power}
Qianli Ma, Jinlong Yang, Siyu Tang, and Michael~J Black.
\newblock The power of points for modeling humans in clothing.
\newblock In \emph{Proceedings of the IEEE/CVF International Conference on
  Computer Vision}, pages 10974--10984, 2021.

\bibitem[Mehta et~al.(2022)Mehta, Chandraker, and Ramamoorthi]{mehta2022level}
Ishit Mehta, Manmohan Chandraker, and Ravi Ramamoorthi.
\newblock A level set theory for neural implicit evolution under explicit
  flows.
\newblock In \emph{Computer Vision--ECCV 2022: 17th European Conference, Tel
  Aviv, Israel, October 23--27, 2022, Proceedings, Part II}, pages 711--729.
  Springer, 2022.

\bibitem[Mildenhall et~al.(2020)Mildenhall, Srinivasan, Tancik, Barron,
  Ramamoorthi, and Ng]{mildenhall2020nerf}
Ben Mildenhall, Pratul~P Srinivasan, Matthew Tancik, Jonathan~T Barron, Ravi
  Ramamoorthi, and Ren Ng.
\newblock Nerf: Representing scenes as neural radiance fields for view
  synthesis.
\newblock In \emph{European conference on computer vision}, pages 405--421.
  Springer, 2020.

\bibitem[Molino et~al.(2003)Molino, Bridson, Teran, and
  Fedkiw]{molino2003crystalline}
Neil Molino, Robert Bridson, Joseph Teran, and Ronald Fedkiw.
\newblock A crystalline, red green strategy for meshing highly deformable
  objects with tetrahedra.
\newblock In \emph{IMR}, pages 103--114. Citeseer, 2003.

\bibitem[Moon et~al.(2022)Moon, Nam, Shiratori, and Lee]{moon20223d}
Gyeongsik Moon, Hyeongjin Nam, Takaaki Shiratori, and Kyoung~Mu Lee.
\newblock 3d clothed human reconstruction in the wild.
\newblock In \emph{Computer Vision--ECCV 2022: 17th European Conference, Tel
  Aviv, Israel, October 23--27, 2022, Proceedings, Part II}, pages 184--200.
  Springer, 2022.

\bibitem[Nam et~al.(2023)Nam, Jung, Oh, and Lee]{nam2023cyclic}
Hyeongjin Nam, Daniel~Sungho Jung, Yeonguk Oh, and Kyoung~Mu Lee.
\newblock Cyclic test-time adaptation on monocular video for 3d human mesh
  reconstruction.
\newblock In \emph{Proceedings of the IEEE/CVF International Conference on
  Computer Vision}, pages 14829--14839, 2023.

\bibitem[Natsume et~al.(2019)Natsume, Saito, Huang, Chen, Ma, Li, and
  Morishima]{natsume2019siclope}
Ryota Natsume, Shunsuke Saito, Zeng Huang, Weikai Chen, Chongyang Ma, Hao Li,
  and Shigeo Morishima.
\newblock Siclope: Silhouette-based clothed people.
\newblock In \emph{Proceedings of the IEEE Conference on Computer Vision and
  Pattern Recognition}, pages 4480--4490, 2019.

\bibitem[Niemeyer et~al.(2020)Niemeyer, Mescheder, Oechsle, and
  Geiger]{niemeyer2020differentiable}
Michael Niemeyer, Lars Mescheder, Michael Oechsle, and Andreas Geiger.
\newblock Differentiable volumetric rendering: Learning implicit 3d
  representations without 3d supervision.
\newblock In \emph{Proceedings of the IEEE/CVF Conference on Computer Vision
  and Pattern Recognition}, pages 3504--3515, 2020.

\bibitem[Omran et~al.(2018)Omran, Lassner, Pons-Moll, Gehler, and
  Schiele]{omran2018neural}
Mohamed Omran, Christoph Lassner, Gerard Pons-Moll, Peter Gehler, and Bernt
  Schiele.
\newblock Neural body fitting: Unifying deep learning and model based human
  pose and shape estimation.
\newblock In \emph{2018 international conference on 3D vision (3DV)}, pages
  484--494. IEEE, 2018.

\bibitem[Onizuka et~al.(2020)Onizuka, Hayirci, Thomas, Sugimoto, Uchiyama, and
  Taniguchi]{onizuka2020tetratsdf}
Hayato Onizuka, Zehra Hayirci, Diego Thomas, Akihiro Sugimoto, Hideaki
  Uchiyama, and Rin-ichiro Taniguchi.
\newblock Tetratsdf: 3d human reconstruction from a single image with a
  tetrahedral outer shell.
\newblock In \emph{Proceedings of the IEEE/CVF Conference on Computer Vision
  and Pattern Recognition}, pages 6011--6020, 2020.

\bibitem[Osher et~al.(2004)Osher, Fedkiw, and Piechor]{osher2004level}
Stanley Osher, Ronald Fedkiw, and K Piechor.
\newblock Level set methods and dynamic implicit surfaces.
\newblock \emph{Appl. Mech. Rev.}, 57\penalty0 (3):\penalty0 B15--B15, 2004.

\bibitem[Pang et~al.(2021)Pang, Chen, Luo, Wu, Yu, and Xu]{pang2021few}
Anqi Pang, Xin Chen, Haimin Luo, Minye Wu, Jingyi Yu, and Lan Xu.
\newblock Few-shot neural human performance rendering from sparse rgbd videos.
\newblock \emph{arXiv preprint arXiv:2107.06505}, 2021.

\bibitem[Park et~al.(2019)Park, Florence, Straub, Newcombe, and
  Lovegrove]{park2019deepsdf}
Jeong~Joon Park, Peter Florence, Julian Straub, Richard Newcombe, and Steven
  Lovegrove.
\newblock Deepsdf: Learning continuous signed distance functions for shape
  representation.
\newblock In \emph{Proceedings of the IEEE/CVF Conference on Computer Vision
  and Pattern Recognition}, pages 165--174, 2019.

\bibitem[Pavlakos et~al.(2018)Pavlakos, Zhu, Zhou, and
  Daniilidis]{pavlakos2018learning}
Georgios Pavlakos, Luyang Zhu, Xiaowei Zhou, and Kostas Daniilidis.
\newblock Learning to estimate 3d human pose and shape from a single color
  image.
\newblock In \emph{Proceedings of the IEEE conference on computer vision and
  pattern recognition}, pages 459--468, 2018.

\bibitem[Pavlakos et~al.(2019)Pavlakos, Choutas, Ghorbani, Bolkart, Osman,
  Tzionas, and Black]{pavlakos2019expressive}
Georgios Pavlakos, Vasileios Choutas, Nima Ghorbani, Timo Bolkart, Ahmed~AA
  Osman, Dimitrios Tzionas, and Michael~J Black.
\newblock Expressive body capture: 3d hands, face, and body from a single
  image.
\newblock In \emph{Proceedings of the IEEE/CVF Conference on Computer Vision
  and Pattern Recognition}, pages 10975--10985, 2019.

\bibitem[Peng et~al.(2021)Peng, Zhang, Xu, Wang, Shuai, Bao, and
  Zhou]{peng2021neural}
Sida Peng, Yuanqing Zhang, Yinghao Xu, Qianqian Wang, Qing Shuai, Hujun Bao,
  and Xiaowei Zhou.
\newblock Neural body: Implicit neural representations with structured latent
  codes for novel view synthesis of dynamic humans.
\newblock In \emph{Proceedings of the IEEE/CVF Conference on Computer Vision
  and Pattern Recognition}, pages 9054--9063, 2021.

\bibitem[Pons-Moll et~al.(2017)Pons-Moll, Pujades, Hu, and
  Black]{pons2017clothcap}
Gerard Pons-Moll, Sergi Pujades, Sonny Hu, and Michael~J Black.
\newblock Clothcap: Seamless 4d clothing capture and retargeting.
\newblock \emph{ACM Transactions on Graphics (ToG)}, 36\penalty0 (4):\penalty0
  1--15, 2017.

\bibitem[Remelli et~al.(2020)Remelli, Lukoianov, Richter, Guillard,
  Bagautdinov, Baque, and Fua]{remelli2020meshsdf}
Edoardo Remelli, Artem Lukoianov, Stephan Richter, Beno{\^\i}t Guillard, Timur
  Bagautdinov, Pierre Baque, and Pascal Fua.
\newblock Meshsdf: Differentiable iso-surface extraction.
\newblock \emph{Advances in Neural Information Processing Systems},
  33:\penalty0 22468--22478, 2020.

\bibitem[RenderPeople()]{renderpeople}
RenderPeople.
\newblock Renderpeople, 2018.

\bibitem[Rosu and Behnke(2023)]{rosu2023permutosdf}
Radu~Alexandru Rosu and Sven Behnke.
\newblock Permutosdf: Fast multi-view reconstruction with implicit surfaces
  using permutohedral lattices.
\newblock In \emph{Proceedings of the IEEE/CVF Conference on Computer Vision
  and Pattern Recognition}, pages 8466--8475, 2023.

\bibitem[Saito et~al.(2019)Saito, Huang, Natsume, Morishima, Kanazawa, and
  Li]{saito2019pifu}
Shunsuke Saito, Zeng Huang, Ryota Natsume, Shigeo Morishima, Angjoo Kanazawa,
  and Hao Li.
\newblock Pifu: Pixel-aligned implicit function for high-resolution clothed
  human digitization.
\newblock In \emph{Proceedings of the IEEE/CVF International Conference on
  Computer Vision}, pages 2304--2314, 2019.

\bibitem[Saito et~al.(2020)Saito, Simon, Saragih, and Joo]{saito2020pifuhd}
Shunsuke Saito, Tomas Simon, Jason Saragih, and Hanbyul Joo.
\newblock Pifuhd: Multi-level pixel-aligned implicit function for
  high-resolution 3d human digitization.
\newblock In \emph{Proceedings of the IEEE/CVF Conference on Computer Vision
  and Pattern Recognition}, pages 84--93, 2020.

\bibitem[Shao et~al.(2022)Shao, Zheng, Zhang, Sun, and
  Liu]{shao2022diffustereo}
Ruizhi Shao, Zerong Zheng, Hongwen Zhang, Jingxiang Sun, and Yebin Liu.
\newblock Diffustereo: High quality human reconstruction via diffusion-based
  stereo using sparse cameras.
\newblock In \emph{Computer Vision--ECCV 2022: 17th European Conference, Tel
  Aviv, Israel, October 23--27, 2022, Proceedings, Part XXXII}, pages 702--720.
  Springer, 2022.

\bibitem[Shen et~al.(2023)Shen, Guo, Kaufmann, Zarate, Valentin, Song, and
  Hilliges]{shen2023x}
Kaiyue Shen, Chen Guo, Manuel Kaufmann, Juan~Jose Zarate, Julien Valentin, Jie
  Song, and Otmar Hilliges.
\newblock X-avatar: Expressive human avatars.
\newblock In \emph{Proceedings of the IEEE/CVF Conference on Computer Vision
  and Pattern Recognition}, pages 16911--16921, 2023.

\bibitem[Shen et~al.(2021)Shen, Gao, Yin, Liu, and Fidler]{shen2021deep}
Tianchang Shen, Jun Gao, Kangxue Yin, Ming-Yu Liu, and Sanja Fidler.
\newblock Deep marching tetrahedra: a hybrid representation for high-resolution
  3d shape synthesis.
\newblock \emph{Advances in Neural Information Processing Systems}, 34, 2021.

\bibitem[Smith et~al.(2019)Smith, Loper, Hu, Mavroidis, and
  Romero]{smith2019facsimile}
David Smith, Matthew Loper, Xiaochen Hu, Paris Mavroidis, and Javier Romero.
\newblock Facsimile: Fast and accurate scans from an image in less than a
  second.
\newblock In \emph{Proceedings of the IEEE/CVF International Conference on
  Computer Vision}, pages 5330--5339, 2019.

\bibitem[Sussman et~al.(1994)Sussman, Smereka, and Osher]{sussman1994level}
Mark Sussman, Peter Smereka, and Stanley Osher.
\newblock A level set approach for computing solutions to incompressible
  two-phase flow.
\newblock \emph{Journal of Computational physics}, 114\penalty0 (1):\penalty0
  146--159, 1994.

\bibitem[Te et~al.(2022)Te, Li, Li, Wang, Hu, and Lu]{te2022neural}
Gusi Te, Xiu Li, Xiao Li, Jinglu Wang, Wei Hu, and Yan Lu.
\newblock Neural capture of animatable 3d human from monocular video.
\newblock In \emph{Computer Vision--ECCV 2022: 17th European Conference, Tel
  Aviv, Israel, October 23--27, 2022, Proceedings, Part VI}, pages 275--291.
  Springer, 2022.

\bibitem[Teran et~al.(2005)Teran, Molino, Fedkiw, and
  Bridson]{teran2005adaptive}
Joseph Teran, Neil Molino, Ronald Fedkiw, and Robert Bridson.
\newblock Adaptive physics based tetrahedral mesh generation using level sets.
\newblock \emph{Engineering with computers}, 21\penalty0 (1):\penalty0 2--18,
  2005.

\bibitem[Tian et~al.(2023)Tian, Zhang, Liu, and Wang]{tian2023recovering}
Yating Tian, Hongwen Zhang, Yebin Liu, and Limin Wang.
\newblock Recovering 3d human mesh from monocular images: A survey.
\newblock \emph{IEEE Transactions on Pattern Analysis and Machine
  Intelligence}, 2023.

\bibitem[Tiwari et~al.(2021)Tiwari, Sarafianos, Tung, and
  Pons-Moll]{tiwari2021neural}
Garvita Tiwari, Nikolaos Sarafianos, Tony Tung, and Gerard Pons-Moll.
\newblock Neural-gif: Neural generalized implicit functions for animating
  people in clothing.
\newblock In \emph{Proceedings of the IEEE/CVF International Conference on
  Computer Vision}, pages 11708--11718, 2021.

\bibitem[Treece et~al.(1999)Treece, Prager, and Gee]{treece1999regularised}
Graham~M Treece, Richard~W Prager, and Andrew~H Gee.
\newblock Regularised marching tetrahedra: improved iso-surface extraction.
\newblock \emph{Computers \& Graphics}, 23\penalty0 (4):\penalty0 583--598,
  1999.

\bibitem[Varol et~al.(2018)Varol, Ceylan, Russell, Yang, Yumer, Laptev, and
  Schmid]{varol2018bodynet}
Gul Varol, Duygu Ceylan, Bryan Russell, Jimei Yang, Ersin Yumer, Ivan Laptev,
  and Cordelia Schmid.
\newblock Bodynet: Volumetric inference of 3d human body shapes.
\newblock In \emph{Proceedings of the European Conference on Computer Vision
  (ECCV)}, pages 20--36, 2018.

\bibitem[Vicini et~al.(2022)Vicini, Speierer, and
  Jakob]{vicini2022differentiable}
Delio Vicini, S{\'e}bastien Speierer, and Wenzel Jakob.
\newblock Differentiable signed distance function rendering.
\newblock \emph{ACM Transactions on Graphics (TOG)}, 41\penalty0 (4):\penalty0
  1--18, 2022.

\bibitem[Wang et~al.(2021)Wang, Mihajlovic, Ma, Geiger, and
  Tang]{wang2021metaavatar}
Shaofei Wang, Marko Mihajlovic, Qianli Ma, Andreas Geiger, and Siyu Tang.
\newblock Metaavatar: Learning animatable clothed human models from few depth
  images.
\newblock \emph{Advances in Neural Information Processing Systems},
  34:\penalty0 2810--2822, 2021.

\bibitem[Wang et~al.(2018)Wang, Liu, Zhu, Tao, Kautz, and
  Catanzaro]{Wang_2018_CVPR}
Ting-Chun Wang, Ming-Yu Liu, Jun-Yan Zhu, Andrew Tao, Jan Kautz, and Bryan
  Catanzaro.
\newblock High-resolution image synthesis and semantic manipulation with
  conditional gans.
\newblock In \emph{Proceedings of the IEEE Conference on Computer Vision and
  Pattern Recognition (CVPR)}, 2018.

\bibitem[Weng et~al.(2022)Weng, Curless, Srinivasan, Barron, and
  Kemelmacher-Shlizerman]{weng2022humannerf}
Chung-Yi Weng, Brian Curless, Pratul~P Srinivasan, Jonathan~T Barron, and Ira
  Kemelmacher-Shlizerman.
\newblock Humannerf: Free-viewpoint rendering of moving people from monocular
  video.
\newblock In \emph{Proceedings of the IEEE/CVF conference on computer vision
  and pattern Recognition}, pages 16210--16220, 2022.

\bibitem[Wu et~al.(2020)Wu, Geng, Zhou, and Fedkiw]{wu2020skinning}
Jane Wu, Zhenglin Geng, Hui Zhou, and Ronald Fedkiw.
\newblock Skinning a parameterization of three-dimensional space for neural
  network cloth.
\newblock \emph{arXiv preprint arXiv:2006.04874}, 2020.

\bibitem[Xiang et~al.(2021)Xiang, Prada, Bagautdinov, Xu, Dong, Wen, Hodgins,
  and Wu]{xiang2021modeling}
Donglai Xiang, Fabian Prada, Timur Bagautdinov, Weipeng Xu, Yuan Dong, He Wen,
  Jessica Hodgins, and Chenglei Wu.
\newblock Modeling clothing as a separate layer for an animatable human avatar.
\newblock \emph{ACM Transactions on Graphics (TOG)}, 40\penalty0 (6):\penalty0
  1--15, 2021.

\bibitem[Xiu et~al.(2022)Xiu, Yang, Tzionas, and Black]{xiu2022icon}
Yuliang Xiu, Jinlong Yang, Dimitrios Tzionas, and Michael~J Black.
\newblock Icon: Implicit clothed humans obtained from normals.
\newblock In \emph{Proceedings of the IEEE/CVF Conference on Computer Vision
  and Pattern Recognition}, pages 13296--13306, 2022.

\bibitem[Xiu et~al.(2023)Xiu, Yang, Cao, Tzionas, and Black]{xiu2023econ}
Yuliang Xiu, Jinlong Yang, Xu Cao, Dimitrios Tzionas, and Michael~J Black.
\newblock Econ: Explicit clothed humans optimized via normal integration.
\newblock In \emph{Proceedings of the IEEE/CVF Conference on Computer Vision
  and Pattern Recognition}, pages 512--523, 2023.

\bibitem[Xue et~al.(2023)Xue, Bhatnagar, Marin, Sarafianos, Xu, Pons-Moll, and
  Tung]{xue2023nsf}
Yuxuan Xue, Bharat~Lal Bhatnagar, Riccardo Marin, Nikolaos Sarafianos, Yuanlu
  Xu, Gerard Pons-Moll, and Tony Tung.
\newblock Nsf: Neural surface fields for human modeling from monocular depth.
\newblock In \emph{ICCV}, 2023.

\bibitem[Yang et~al.(2018)Yang, Ouyang, Wang, Ren, Li, and Wang]{yang20183d}
Wei Yang, Wanli Ouyang, Xiaolong Wang, Jimmy Ren, Hongsheng Li, and Xiaogang
  Wang.
\newblock 3d human pose estimation in the wild by adversarial learning.
\newblock In \emph{Proceedings of the IEEE Conference on Computer Vision and
  Pattern Recognition}, pages 5255--5264, 2018.

\bibitem[Yariv et~al.(2020)Yariv, Kasten, Moran, Galun, Atzmon, Ronen, and
  Lipman]{yariv2020multiview}
Lior Yariv, Yoni Kasten, Dror Moran, Meirav Galun, Matan Atzmon, Basri Ronen,
  and Yaron Lipman.
\newblock Multiview neural surface reconstruction by disentangling geometry and
  appearance.
\newblock \emph{Advances in Neural Information Processing Systems},
  33:\penalty0 2492--2502, 2020.

\bibitem[Yu et~al.(2019)Yu, Zheng, Zhong, Zhao, Dai, Pons-Moll, and
  Liu]{yu2019simulcap}
Tao Yu, Zerong Zheng, Yuan Zhong, Jianhui Zhao, Qionghai Dai, Gerard Pons-Moll,
  and Yebin Liu.
\newblock Simulcap: Single-view human performance capture with cloth
  simulation.
\newblock In \emph{Proceedings of the IEEE Conference on Computer Vision and
  Pattern Recognition}, 2019.

\bibitem[Yu et~al.(2023)Yu, Cheng, Liu, Wu, and Lin]{yu2023monohuman}
Zhengming Yu, Wei Cheng, Xian Liu, Wayne Wu, and Kwan-Yee Lin.
\newblock Monohuman: Animatable human neural field from monocular video.
\newblock In \emph{Proceedings of the IEEE/CVF Conference on Computer Vision
  and Pattern Recognition}, pages 16943--16953, 2023.

\bibitem[Zanfir et~al.(2020)Zanfir, Bazavan, Xu, Freeman, Sukthankar, and
  Sminchisescu]{zanfir2020weakly}
Andrei Zanfir, Eduard~Gabriel Bazavan, Hongyi Xu, William~T Freeman, Rahul
  Sukthankar, and Cristian Sminchisescu.
\newblock Weakly supervised 3d human pose and shape reconstruction with
  normalizing flows.
\newblock In \emph{European Conference on Computer Vision}, pages 465--481.
  Springer, 2020.

\bibitem[Zhang et~al.(2023)Zhang, Sun, Yang, Chen, and Yang]{zhang2023global}
Zechuan Zhang, Li Sun, Zongxin Yang, Ling Chen, and Yi Yang.
\newblock Global-correlated 3d-decoupling transformer for clothed avatar
  reconstruction.
\newblock \emph{arXiv preprint arXiv:2309.13524}, 2023.

\bibitem[Zhao et~al.(2020)Zhao, Sun, Zhang, Tang, and Qian]{zhao2020monocular}
Chaoqiang Zhao, Qiyu Sun, Chongzhen Zhang, Yang Tang, and Feng Qian.
\newblock Monocular depth estimation based on deep learning: An overview.
\newblock \emph{Science China Technological Sciences}, 63\penalty0
  (9):\penalty0 1612--1627, 2020.

\bibitem[Zhao et~al.(2022)Zhao, Yang, Zhang, Lin, Zhang, Yu, and
  Xu]{zhao2022humannerf}
Fuqiang Zhao, Wei Yang, Jiakai Zhang, Pei Lin, Yingliang Zhang, Jingyi Yu, and
  Lan Xu.
\newblock Humannerf: Efficiently generated human radiance field from sparse
  inputs.
\newblock In \emph{Proceedings of the IEEE/CVF Conference on Computer Vision
  and Pattern Recognition}, pages 7743--7753, 2022.

\bibitem[Zhao et~al.(1996)Zhao, Chan, Merriman, and Osher]{zhao1996variational}
Hong-Kai Zhao, Tony Chan, Barry Merriman, and Stanley Osher.
\newblock A variational level set approach to multiphase motion.
\newblock \emph{Journal of computational physics}, 127\penalty0 (1):\penalty0
  179--195, 1996.

\bibitem[Zheng et~al.(2023{\natexlab{a}})Zheng, Li, Wang, and
  Yu]{zheng2023learning}
Ruichen Zheng, Peng Li, Haoqian Wang, and Tao Yu.
\newblock Learning visibility field for detailed 3d human reconstruction and
  relighting.
\newblock In \emph{Proceedings of the IEEE/CVF Conference on Computer Vision
  and Pattern Recognition}, pages 216--226, 2023{\natexlab{a}}.

\bibitem[Zheng et~al.(2019)Zheng, Yu, Wei, Dai, and Liu]{zheng2019deephuman}
Zerong Zheng, Tao Yu, Yixuan Wei, Qionghai Dai, and Yebin Liu.
\newblock Deephuman: 3d human reconstruction from a single image.
\newblock In \emph{Proceedings of the IEEE/CVF International Conference on
  Computer Vision}, pages 7739--7749, 2019.

\bibitem[Zheng et~al.(2021)Zheng, Yu, Liu, and Dai]{zheng2021pamir}
Zerong Zheng, Tao Yu, Yebin Liu, and Qionghai Dai.
\newblock Pamir: Parametric model-conditioned implicit representation for
  image-based human reconstruction.
\newblock \emph{IEEE transactions on pattern analysis and machine
  intelligence}, 2021.

\bibitem[Zheng et~al.(2023{\natexlab{b}})Zheng, Zhao, Zhang, Liu, and
  Liu]{zheng2023avatarrex}
Zerong Zheng, Xiaochen Zhao, Hongwen Zhang, Boning Liu, and Yebin Liu.
\newblock Avatarrex: Real-time expressive full-body avatars.
\newblock \emph{arXiv preprint arXiv:2305.04789}, 2023{\natexlab{b}}.

\bibitem[Zhou et~al.(2022)Zhou, Yu, Shao, and Li]{zhou2022hdhuman}
Tiansong Zhou, Tao Yu, Ruizhi Shao, and Kun Li.
\newblock Hdhuman: High-quality human performance capture with sparse views.
\newblock \emph{CoRR}, 2022.

\end{thebibliography}
}
\end{document}


\maketitle

\appendix

\section{Ray-Tracing the Implicit Surface Directly}
As an alternative to Marching Tetrahedra, consider casting a ray to find an intersection point with the implicit surface and subsequently using the normal vector defined (directly) by the implicit surface at that intersection point.
A number of existing works consider such approaches in various ways, see e.g.\ \cite{niemeyer2020differentiable,yariv2020multiview,bangaru2022differentiable,chen2022gdna,vicini2022differentiable}.
Perturbations of the intersection point depend on perturbations of the $\phi$ values on the vertices of the tetrahedron that the intersection point lies within.
If a change in $\phi$ values causes the intersection point to no longer be contained inside the tetrahedron, one would need to discontinuously jump to some other tetrahedron (which could be quite far away, if it even exists).
A potential remedy for this would be to define a virtual implicit surface that extends out of the tetrahedron in a way that provides some sort of continuity (especially along silhouette boundaries).

Comparatively, our Marching Tetrahedra approach allows us to presume (for example) that the point of intersection remains fixed on the face of the triangle even as the triangle moves.
Since the implicit surface has no explicit parameterization, one is unable to similarly hold the intersection point fixed.
The implicit surface utilizes an Eulerian point of view where the rays (which represent the discretization) are held fixed while the implicit surface moves (as $\phi$ values change), in contrast to our Lagrangian discretization where the rays are allowed to move/bend in order to follow fixed intersection points during differentiation.
A similar approach for an implicit surface would hold the intersection point inside the tetrahedron fixed even as $\phi$ changes.
Although such an approach holds potential due to the fact that implicit surfaces are amenable to computing derivatives off of the surface itself, the merging/pinching of isocontours created by convexity/concavity would likely lead to various difficulties.
Furthermore, other issues would need to be addressed as well, e.g.\ the gradients (and thus normals) are only piecewise constant (and thus discontinuous) in the piecewise linear tetrahedral mesh basis.

\section{Skinning}
There are two options for the algorithm ordering between skinning and Marching Tetrahedra (the latter of which reverses the order in Figure \ref{fig:pipeline}).
For skinning the triangle mesh, the skinned position of each triangle mesh vertex is $v_i(\theta,\phi) = \sum_j w_{ij}(\phi)T_j(\theta)v_i^j(\phi)$ where $v_i^j$ is the location of $v_i$ in the untransformed reference space of joint $j$.
Unlike in Section \ref{sec:skinning} where $w_{kj}$ and $u_k^j$ were fixed, $w_{ij}$ and $v_i^j$ both vary yielding three terms in the product rule.
$\partial v_{i}^j/\partial\phi$ is computed according to Equation \ref{eq:mt_grad}, noting that $u_{k_1}$ and $u_{k_2}$ are fixed.
$w_{ij}(\phi)$ is defined similarly to Equation \ref{eq:zero_crossing},
\begin{equation}\label{eq:weights_crossing}
    w_{ij} = \frac{-\phi_{k_2}}{\phi_{k_1} - \phi_{k_2}}w_{k_1j} + \frac{\phi_{k_1}}{\phi_{k_1} - \phi_{k_2}}w_{k_2j}
\end{equation}
where $w_{k_1 j}$ and $w_{k_2 j}$ are fixed; similar to Equation \ref{eq:mt_grad}, $\partial w_{ij} / \partial \phi$ will contain $\mathcal{O}(1/\epsilon)$ coefficients.
For skinning the tetrahedral mesh, Equations \ref{eq:zero_crossing} and \ref{eq:mt_grad} directly define $v_i$ and $\partial v_i/\partial\phi$ since the skinning is moved to the tetrahedral mesh vertices $u_k$.
Then, $\partial v_i/ \partial u_k$ is computed according to Equation \ref{eq:zero_crossing} in order to chain rule to skinning (i.e.\ to $\partial u_k/ \partial \theta$, which is computed according to the equations in Section \ref{sec:skinning}).

\section{Image Rasterization Implementation}
\subsection{Normals}\label{sec:geometry}
Recall (from Section \ref{sec:mt}) that triangle vertices are reordered (if necessary) in order to obtain outward-pointing face normals.
The area-weighted outward face normal is
\begin{equation}\label{eq:face_normal}
    n_f(v_1,v_2,v_3) =\frac 12 (v_2-v_1) \times (v_3-v_1)
\end{equation}
where
\begin{equation}\label{eq:area_3d}
    Area(v_1,v_2,v_3) = \frac 12||(v_2-v_1) \times (v_3-v_1)||_2
\end{equation}
is the area weighting.
Area-averaged vertex unit normals $\hat{n}_v$ are computed via
\begin{align}\label{eq:vert_normal}
\begin{split}
     n_v = \sum_f n_f \quad \quad  \quad \quad 
     \hat{n}_v &= \frac{n_v}{||n_v||_2}
\end{split}
\end{align}
where $f$ ranges over all the triangle faces that include vertex $v$. Note that one can drop the 1/2 in Equation \ref{eq:face_normal}, since it cancels out when computing $\hat{n}_v$ in Equation \ref{eq:vert_normal}.

\subsection{Camera Model}\label{sec:camera_model}
\begin{figure}[ht]
    \centering
    \includegraphics[width=\linewidth]{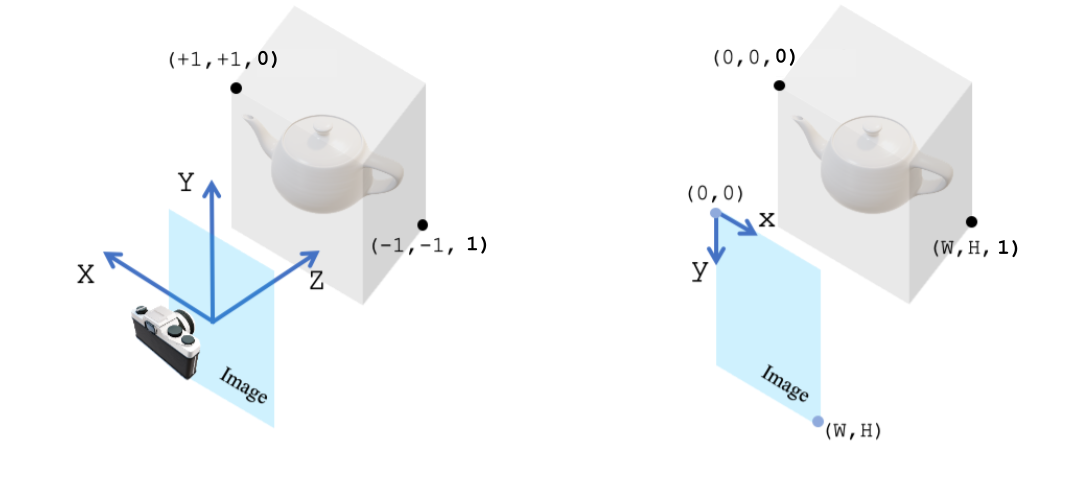}
    \caption{The normalized device (left) and screen (right) coordinate systems used during rasterization (based on Pytorch3D conventions).}
    \label{fig:coord_systems}
\end{figure}

The camera rotation and translation are used to transform each vertex $v_g$ of the geometry to the camera view coordinate system (where the origin is located at the camera aperture), i.e.\ $v_{c} = Rv_g + T$.
The normalized device coordinate system normalizes geometry in the viewing frustum (with $z\in [n, f]$) so that all $x,y \in [-1,1]$ and all $z \in [0,1]$.
See Figure \ref{fig:coord_systems}, left.
Vertices are transformed into this coordinate system via
\begin{align}\label{eq:projection_matrix}
\begin{split}
    \begin{bmatrix}[v_{NDC}]\hspace{1mm} z_c\\ z_c\end{bmatrix} &=  \begin{bmatrix}
    \frac{2n}{W} & 0 & 0  & 0 \\
    0 & \frac{2n}{H} & 0 & 0\\
    0 & 0 & \frac{f}{f-n} & \frac{-fn}{f-n} \\
    0 & 0 & 1 & 0 \end{bmatrix}\begin{bmatrix}[v_c]\\1 \end{bmatrix}
\end{split}
\end{align}
where $H = 2n\tan(\theta_{\text{fov}}/2)$ is the height of the image, $\theta_{\text{fov}}$ is the field of view, $W=Ha$ is the width of the image, and $a$ is the aspect ratio.
The screen coordinate system is obtained by transforming the origin to the top left corner of the image, with $+x$ pointing right and $+y$ pointing down.
See Figure \ref{fig:coord_systems}, right.
Vertices are transformed into this coordinate system via
\begin{equation}\label{eq:screen_proj}
    \begin{bmatrix}[v']\\1 \end{bmatrix} = \begin{bmatrix}
    -W/2 & 0 & 0 & W/2 \\
    0 & -H/2 & 0 & H/2\\
    0 & 0 & 1 & 0\\
    0 & 0 & 0 & 1 \\
    \end{bmatrix}\begin{bmatrix}[v_{NDC}]\\1 \end{bmatrix}
\end{equation}

or via
\begin{align}\label{eq:full_transform}
\begin{split}
    \begin{bmatrix}[v']\hspace{1mm} z_c\\ z_c\end{bmatrix} &=
    \begin{bmatrix}
    -n & 0 & W/2 & 0\\
    0 & -n & H/2 & 0\\
    0 & 0 & \frac{f}{f-n} & \frac{-fn}{f-n} \\
    0 & 0 & 1 & 0\\
    \end{bmatrix}
     \begin{bmatrix}[v_c]\\1 \end{bmatrix}\\
\end{split}
\end{align}
which is obtained by multiplying both sides of Equation \ref{eq:screen_proj} by $z_c$ and substituting in Equation \ref{eq:projection_matrix}.

\subsection{Normal Map}
For each pixel, a ray is cast from the camera aperture through the pixel center to find its first intersection with the triangulated surface at a point $p$ in world space.
Denoting $v_1, v_2, v_3$ as the vertices of the intersected triangle, barycentric weights for the intersection point
\begin{align}\label{eq:ray_trace}
\begin{split}
    \hat{\alpha}_1 &= \frac{Area(p, v_2, v_3)}{Area(v_1, v_2, v_3)} \\
    \hat{\alpha}_2 &= \frac{Area(v_1, p, v_3)}{Area(v_1, v_2, v_3)}  \\
    \hat{\alpha}_3 &= \frac{Area(v_1, v_2,p)}{Area(v_1, v_2, v_3)}  
\end{split}
\end{align}
are used to compute a rotated (into screen space) unit normal from the unrotated vertex unit normals (see Equation \ref{eq:vert_normal}) via
\begin{equation}\label{eq:normals}
    \hat{n} = R \frac{\hat{\alpha}_1 \hat{n}_{v_1} + \hat{\alpha}_2 \hat{n}_{v_2} + \hat{\alpha}_3 \hat{n}_{v_3}}{||\hat{\alpha}_1 \hat{n}_{v_1} + \hat{\alpha}_2 \hat{n}_{v_2} + \hat{\alpha}_3 \hat{n}_{v_3}||}
\end{equation}
for the normal map.
Note that dropping the denominators in Equation \ref{eq:ray_trace} does not change $\hat{n}$.

\subsection{Scanline Rendering}
After projecting a visible triangle into the screen coordinate system (via Equation \ref{eq:full_transform}), its projected area can be computed as
\begin{align}\label{eq:area_2d}
\begin{split}
    Area2D(v_1', v_2', v_3') &= -\frac 12det\begin{pmatrix}
    x_{2}'-x_{1}' & y_{2}'-y_{1}' \\
    x_{3}'-x_{1}' & y_{3}'-y_{1}' \\
    \end{pmatrix}
\end{split}
\end{align}
similar to Equation \ref{eq:area_3d} (where the negative sign accounts for the fact that visible triangles have normals pointing towards the camera).
When a projected triangle overlaps a pixel center $p'$, barycentric weights for $p'$ are computed by using $Area2D$ instead of $Area$ in Equation \ref{eq:ray_trace}.
Notably, un-normalized world space barycentric weights can be computed from un-normalized screen space barycentric weights via $\alpha_1 = z_2' z_3' \alpha_1'$, $\alpha_2 = z_1' z_3' \alpha_2'$, $\alpha_3 = z_1' z_2'\alpha_3'$ or
\begin{align}\label{eq:alphas}
\begin{split}
    \alpha_1 &= z_2' z_3' Area2D(p', v_2', v_3') \\
    \alpha_2 &= z_1' z_3' Area2D(v_1', p', v_3') \\
    \alpha_3 &= z_1' z_2' Area2D(v_1', v_2', p') 
\end{split}
\end{align}
giving 
\begin{equation}\label{eq:new_normals}
    \hat{n} = R\frac{\alpha_1 \hat{n}_{v_1} + \alpha_2 \hat{n}_{v_2} + \alpha_3 \hat{n}_{v_3}}{||\alpha_1 \hat{n}_{v_1} + \alpha_2 \hat{n}_{v_2} + \alpha_3 \hat{n}_{v_3}||}
\end{equation}
as an (efficient) alternative to Equation \ref{eq:normals}.
If more than one triangle overlaps $p'$, the closest one (i.e.\ the one with the smallest value of $z' = \hat{\alpha}_1'z_1'+\hat{\alpha}_2'z_2'+\hat{\alpha}_3'z_3'$ at $p'$) is chosen.

\subsection{Computing Gradients}
For each pixel overlapped by the triangle mesh, the derivative of the normal (in Equation \ref{eq:new_normals}) with respect to the vertices of the triangle mesh is required, i.e.\ $\partial \alpha_i / \partial v_g$ and $\partial \hat{n}_{v_i} / \partial v_g$ are required.
$\partial \alpha_i / \partial v'$ can be computed from Equations \ref{eq:alphas} and \ref{eq:area_2d}, $\partial v'/\partial v_c$ can be computed from Equation \ref{eq:full_transform}, and $\partial v_c/\partial v_g$ can be computed from $v_{c} = Rv_g + T$.
$\partial \hat{n}_{v_i} / \partial v_g$ can be computed from Equations \ref{eq:vert_normal} and \ref{eq:face_normal}.

\section{Motion Sequence}\label{sec:pose_generalization}
We first evaluate how well our method generalizes to different poses of the same person using motion sequence data from RenderPeople \cite{renderpeople}.
In each experiment, shown in Figure \ref{fig:pose_interp}, we use every $k$th frame of the motion sequence as the training dataset, where $k \in \{2,4,6,8\}$.
To generate training data for each frame in the motion sequence, the person is rendered from a fixed camera view and the body pose is estimated by fitting the SMPL skeleton to OpenPose keypoints \cite{cao2019openpose} predicted from multiple views of the scan (as in \cite{zheng2021pamir}).
The trained models are then evaluated on unseen intermediate frames in the test dataset. We observe that the more frames our method is trained on, the lower the test set generalization error (although using more examples may lead to greater instability during training).

\begin{figure}[h]
\begin{subfigure}[b]{0.19\linewidth}
    \includegraphics[width=\linewidth]{figures/experiments/pose_generalization/test_8/pred_2.png}
    \includegraphics[width=\linewidth]{figures/experiments/pose_generalization/test_8/pred_5.png}
    \includegraphics[width=\linewidth]{figures/experiments/pose_generalization/test_8/pred_6.png}
    \includegraphics[width=\linewidth]{figures/experiments/pose_generalization/test_8/pred_7.png}
    \caption{$k=8$}
\end{subfigure}
\begin{subfigure}[b]{0.19\linewidth}
   \includegraphics[width=\linewidth]{figures/experiments/pose_generalization/test_6/pred_2.png}
    \includegraphics[width=\linewidth]{figures/experiments/pose_generalization/test_6/pred_5.png}
    \includegraphics[width=\linewidth]{figures/experiments/pose_generalization/test_6/pred_6.png}
    \includegraphics[width=\linewidth]{figures/experiments/pose_generalization/test_6/pred_7.png}
    \caption{$k=6$}
\end{subfigure}
\begin{subfigure}[b]{0.19\linewidth}
   \includegraphics[width=\linewidth]{figures/experiments/pose_generalization/test_4/pred_2.png}
    \includegraphics[width=\linewidth]{figures/experiments/pose_generalization/test_4/pred_5.png}
    \includegraphics[width=\linewidth]{figures/experiments/pose_generalization/test_4/pred_6.png}
    \includegraphics[width=\linewidth]{figures/experiments/pose_generalization/test_4/pred_7.png}
    \caption{$k=4$}
\end{subfigure}
\begin{subfigure}[b]{0.19\linewidth}
    \includegraphics[width=\linewidth]{figures/experiments/pose_generalization/test_2/pred_2.png}
    \includegraphics[width=\linewidth]{figures/experiments/pose_generalization/test_2/pred_5.png}
    \includegraphics[width=\linewidth]{figures/experiments/pose_generalization/test_2/pred_6.png}
    \includegraphics[width=\linewidth]{figures/experiments/pose_generalization/test_2/pred_7.png}
    \caption{$k=2$}
\end{subfigure}
\begin{subfigure}[b]{0.19\linewidth}
    \includegraphics[width=\linewidth]{figures/experiments/pose_generalization/gt/gt_2.png}
    \includegraphics[width=\linewidth]{figures/experiments/pose_generalization/gt/gt_5.png}
    \includegraphics[width=\linewidth]{figures/experiments/pose_generalization/gt/gt_6.png}
    \includegraphics[width=\linewidth]{figures/experiments/pose_generalization/gt/gt_7.png}
    \caption{GT}
\end{subfigure}
\caption{Models trained on a decreasing interval of motion sequence frames (inference from novel poses/frames is shown). The first four columns correspond to models trained on every 8th, 6th, 4th, and 2nd frame, and the last column shows the ground truth normal maps.}
\label{fig:pose_interp}
\end{figure}

{
\small
\bibliographystyle{ieeenat_fullname}
\bibliography{mainbib}
}